\documentclass[pdflatex,sn-mathphys-num]{sn-jnl}

\usepackage{graphicx}%
\usepackage{multirow}%
\usepackage{amsmath,amssymb,amsfonts}%
\usepackage{amsthm}%
\usepackage{mathrsfs}%
\usepackage[title]{appendix}%
\usepackage{xcolor}%
\usepackage{textcomp}%
\usepackage{manyfoot}%
\usepackage{booktabs}%
\usepackage{algorithm}%
\usepackage{algorithmicx}%
\usepackage{algpseudocode}%
\usepackage{listings}%
\usepackage{pdflscape}
\usepackage{longtable}
\usepackage{bbm}
\usepackage[nameinlink]{cleveref}
\crefname{figure}{Fig.}{Figs.}
\Crefname{figure}{Figure}{Figures}
\crefname{table}{Table}{Tables}
\Crefname{table}{Table}{Tables}
\usepackage{url}
\usepackage{siunitx}
\usepackage{placeins}
\usepackage{minitoc}
\mtcsettitle{minilof}{Supplementary Figures}
\mtcsettitle{minilot}{Supplementary Tables}

\theoremstyle{thmstyleone}%
\theoremstyle{thmstyletwo}%

\theoremstyle{thmstylethree}%

\begin{document}

\title[Clustering-Based ART Ensembles for RF Approximation]{Clustering-Based Artificial Tree Ensembles for Random Forest Approximation}


\author*[1]{\fnm{Lea L.} \sur{Kronziel}}\email{l.kronziel@uni-luebeck.de}

\author[1]{\fnm{Maren S.} \sur{Wieder}}\email{m.wieder@gy-mi.de}

\author[1]{\fnm{Silke} \sur{Szymczak}}\email{silke.szymczak@uni-luebeck.de}

\author[1,2,3]{\fnm{Björn-Hergen} \sur{Laabs}}\email{bjoern-hergen.laabs@med.uni-goettingen.de}

\affil[1]{\orgdiv{Institute of Medical Biometry and Statistics}, \orgname{University of Luebeck}, \orgaddress{\street{Ratzeburger Allee 160}, \city{L\"ubeck}, \postcode{23562}, \country{Germany}}}

\affil[2]{\orgdiv{Department of Medical Statistics}, \orgname{University Medical Center G\"ottingen}, \orgaddress{\street{Humboldtallee 32}, \city{Göttingen}, \postcode{37073}, \state{Lower Saxony}, \country{Germany}}}

\affil[3]{\orgdiv{CAIMed}, \orgname{Lower Saxony Center for Artificial Intelligence and Causal Methods in Medicine}, \orgaddress{\state{Lower Saxony}, \country{Germany}}}



\abstract{Random Forests (RFs) offer strong prediction performance but are difficult to interpret. 
Most representative trees (MRTs) and artificial representative trees (ARTs) provide surrogate explanations, but a single tree may miss patterns within a forest.
We investigate benefits and limitations of clustering-based representative tree ensembles for RF surrogate modeling and assess the contribution of similarity-based clustering relative to random tree partitions.
Using linear and nonlinear simulations and regression benchmarks, we evaluate fidelity, prediction accuracy, complexity, and prediction stability against equally sized ensembles of top-ranked MRTs and randomly selected trees.
Increasing ensemble size generally improves fidelity, accuracy, and prediction stability. 
ART ensembles show higher fidelity in simulations and often greater prediction stability for small ensembles.
Compared with random tree partitions, similarity-based clustering generally improves ART fidelity in simulations and often on benchmarks.
On benchmark datasets, MRT-based ensembles generally achieve higher fidelity and prediction accuracy, whereas ART ensembles have fewer terminal nodes. ART ensembles provide compact surrogates when reduced complexity matters and their fidelity and prediction accuracy meet application requirements. These findings clarify both the trade-offs of representative tree ensembles as RF surrogates and the contribution of similarity-based clustering beyond random clustering.
}



\maketitle
\section{Introduction}
Machine learning methods like Random Forests (RFs) are popular for their robustness and ability to handle complex datasets. 
In particular, RF models are known for their excellent prediction performance.  
However, their interpretability is limited due to the large number of trees, which makes it challenging to understand the structure of the model.

Most representative trees (MRTs) have the potential to act as surrogate models by selecting a tree that represents the main structure of the forest \cite{banerjee_identifying_2012}.
Alternatively, artificial representative trees (ARTs), trees that closely resemble the RF without being part of the initial ensemble, can be used to increase the similarity of the predictions and improve interpretability \cite{laabs_construction_2024}.
However, a single representative tree may not capture the full spectrum of different patterns and complex interactions within an RF, resulting in an oversimplified explanation.

To overcome this limitation, it is reasonable to focus on sets of representative trees instead of relying on only one tree. An intuitive approach to gaining a more comprehensive understanding of the fundamental structure of RFs is to exploit the heterogeneity of trees within an RF by grouping them into clusters \cite{sies_c443_2020}. An MRT can be defined for each cluster, yielding a set of MRTs.

Clustering-based approaches for analyzing ensembles of decision trees are not new. Previous work has used clustering to explore heterogeneity between trees or to improve visualization and understanding of the structure of the ensemble \cite{sies_c443_2020}. 
However, these approaches primarily focus on descriptive insights and do not systematically evaluate the prediction performance of the resulting ensembles nor the role of clustering for constructing surrogate models of RF. 
In particular, it remains unclear how such ensembles balance fidelity to the RF, prediction performance, and model complexity, and under which conditions their reduced complexity provides practical benefits.

Consequently, we conducted an extensive simulation study and benchmark application to compare clustering-based ensembles as RF surrogates with respect to fidelity, prediction accuracy, model complexity, and prediction stability.
We extend single-tree ART representation by constructing one ART per cluster of RF trees. Comparisons with top-ranked MRTs and randomly selected RF trees at matched ensemble sizes provide additional baselines, while comparisons within identical clusters assess ART construction relative to MRT selection.
We evaluate these approaches in linear and nonlinear simulation scenarios and nineteen benchmark datasets, with ART comparisons restricted to datasets with complete results. 
We additionally assess prediction stability across repetitions or cross-validation folds. 
In the simulations, sensitivity analyses examine variable preselection.
In both simulations and benchmark experiments, we assess the contribution of similarity-based tree clustering by comparing WSV-based and prediction-based clustering with random tree clustering at matched ensemble sizes.

Our contribution is a systematic evaluation of representative tree ensembles as RF surrogates, examining both their trade-offs between performance and complexity and the additional value of tree clustering relative to random tree partitions.

\section{Background}
\subsection{Random Forests}
An RF is a supervised machine learning algorithm for classification, regression, and survival analysis \cite{breiman_random_2001,ishwaran_random_2008}.
RFs are an ensemble of decision trees, where each tree is trained on a different subset of the training data obtained by bootstrapping or subsampling. 
The trees are constructed using recursive binary splits, and at each split only a random subset of features (of size \textit{mtry}) is considered to determine the best split using a criterion based on impurity reduction, such as the Gini impurity.
By aggregating the predictions of multiple trees, RFs reduce overfitting and improve the accuracy and robustness of the model compared to individual decision trees \cite{breiman_random_2001}. 
However, due to the complexity of the ensemble, RF models are difficult to interpret and thus are often considered "black boxes".
Although RFs can provide measures of variable importance, they only offer insight into the impact of variables on the model's predictions. They do not capture potential interactions between variables or reflect the underlying structure of individual decision paths within the ensemble.
Surrogate models can be used for this purpose.
These are easy-to-interpret models, such as linear regression or decision trees, that should resemble the RF as closely as possible so that they can be used instead of the RF for interpretation. In order to achieve the greatest possible similarity between the surrogate model and the RF, the surrogate model can be trained on the predictions of the RF. However, such approaches primarily focus on reproducing the predictions of the RF and may not preserve its underlying structural characteristics. As an alternative, representative trees aim to approximate the RF using splitting rules observed in the ensemble, thereby reflecting its structure more directly while remaining close to its predictions.

\subsection{Most Representative Trees}
To improve the interpretability of RFs, an MRT can be selected from the ensemble \cite{banerjee_identifying_2012}. This approach has the advantage that a component of the RF is used directly.
The MRT is the tree that best represents the entire forest and can be used as a surrogate model to interpret the RF's predictions. 
The pairwise distances between the trees in the ensemble are determined, and the tree with the lowest average distance to all others is selected. Distance metrics include the similarity of predictions or the used splitting variables (SV) \cite{banerjee_identifying_2012}. 
The weighted SV (WSV) metric extends the SV measure by additionally accounting for the depth at which variables are used for splitting within a tree \cite{laabs_identification_2024}. For each tree $T_i$, a usage score is computed for all variables $v \in V$ of the training data, defined as
\begin{align}
	U_i(v) = \frac{\sum_{k=1}^{K_i} \frac{\mathbbm{1}_{[s_i(k) = v]}}{2^{\,l_i(k) - 1}}}{\max_k\bigl(l_i(k)\bigr)},
\end{align}
where $\mathbbm{1}_{[s_i(k) = v]}$ is an indicator function that equals 1 if variable $v$ is used as the splitting variable at node $k$ of tree $T_i$, and 0 otherwise. The term $l_i(k)$ denotes the depth of node $k$, which serves as a weight, assigning higher importance to splits closer to the root of the tree. 

The WSV distance between two trees $T_i$ and $T_j$ is then defined as
\begin{align}
	d_{\mathrm{WSV}}(T_i, T_j) = \sum_{v \in V} \left( U_i(v) - U_j(v) \right)^2.
\end{align}

\subsection{Artificial Representative Trees}
However, due to the random selection of candidate variables at each split in a single tree, it cannot be guaranteed that the first split contains an effect variable. This can lead to incorrect conclusions when interpreting the trees. 
Therefore, ARTs were proposed as a new type of surrogate model \cite{laabs_construction_2024}. 
The ART is built by combining splits that occur in the RF trees, ensuring that the newly created tree closely resembles the structure of the trees in the RF. Similar to the MRT, different distance metrics can be used to optimize the ART. 
Additionally, the ART can be limited to splits involving only the most important variables using the hyperparameter \emph{imp.num.var}. This parameter can either be set to a fixed numerical value, specifying how many of the top-ranked variables should be considered based on a chosen variable importance measure of the RF, or it can be determined automatically.
In the latter case, the \emph{Boruta} variable selection algorithm can be used to identify a subset of relevant variables that have been identified as important based on statistical tests \cite{kursa_feature_2010}. This flexibility allows the ART to focus on the most influential variables while reducing the inclusion of less informative or noisy features.
For continuous variables, the splits can also be restricted to user defined quantiles (hyperparameter $probs\_quantiles$). Instead of considering all possible split points, only selected quantiles, such as the 25\%, 50\%, and 75\% quantiles, are used. This reduces the number of candidate splits and thereby limits the space of possible ARTs. As a result, the algorithm becomes more efficient, as the optimal ART can be identified more quickly.
Similar to the RF algorithm, the tree growth of a node can be controlled by specifying a minimum number of observations per terminal node (hyperparameter $min.bucket$). In practice, this means that a node is only split further if each resulting child node would still contain at least min.bucket observations. If this condition is not met, the splitting process stops and the node becomes a terminal node.

\section{Related work}

Several approaches have been proposed to combine decision trees or RF and clustering techniques to improve model performance. For example, in one approach clustering is used as preprocessing step \cite{parisot_using_2013}. 
Specifically, the training data is clustered using algorithms such as k-means. The resulting cluster assignments are then added as an additional categorical feature to the dataset. 
This extended dataset is then used to train a decision tree. Consequently, the tree structure can represent all clusters simultaneously. 
The main advantage of this method is that the complete structural information of the clustering step is preserved and incorporated into the decision-making process. 
However, this can lead to very deep and complex tree structures, reducing interpretability. 
Furthermore, the inherent instability of decision trees remains a limitation.
In another approach, clustering techniques are combined with RF by focusing on the heterogeneity among the trees of the ensemble \cite{sies_c443_2020}.
The trees are clustered using the Partitioning Around Medoids (PAM) method, based on a defined similarity measure. One tree is then selected per cluster, for example, based on the highest average similarity to the other trees in the cluster.
The aim is to explore the heterogeneity within an ensemble of trees. Although this method provides insight into the structure of the RF, it does not directly address the interpretability of the ensemble.
It also does not evaluate the suitability of representative trees as surrogate models or the predictive performance of those trees.
Beyond clustering-based approaches, several methods aim to extract a small ensemble of rules as simplified and more interpretable representations from tree ensembles. The interpretable trees (\emph{inTrees}) framework extracts, prunes, and selects rules from tree ensembles such as RFs and combines them into a compact predictive model \cite{deng_interpreting_2019}. Similarly, the Stable and Interpretable RUle Set (\emph{SIRUS}) method constructs a stable and sparse rule-based model by selecting frequently occurring decision rules across the ensemble \cite{benard_interpretable_2021}. While these approaches improve interpretability by operating on the level of decision rules, they do not preserve the original tree structure and therefore provide a different perspective on the model than tree-based surrogates.
Other approaches focus on using small sets of trees as surrogate models. For instance, a small ensembles of decision trees or most representative trees is used to approximate the behavior of RFs \cite{szepannek_cant_2024}. These so-called groves aim to balance model complexity and explanatory power by combining multiple simple trees. However, these methods do not explicitly exploit clustering of trees within the forest to guide the construction of such ensembles.

Beyond RF surrogate modeling, recent work has investigated how feature selection can guide the construction of ensemble members. 
\citet{jiang_neighborhood_2025} use neighborhood rough sets to generate feature subsets, while \citet{jiang_ensemble_2026} employ neighborhood granular combination entropy for this purpose. 
Both approaches aim to construct accurate and diverse ensembles for software defect prediction. 
However, they introduce diversity through feature-space manipulation when training prediction models, whereas we exploit differences among the trees of an existing RF through clustering.

Clustering and model combination also play a role in fuzzy modeling, where they support representative feature selection and forecast combination.
\citet{tak_meta_2021} combines forecasts from single-hidden-layer feed-forward neural networks trained with different random initializations, using meta fuzzy functions to assign different weights rather than averaging the forecasts equally. 
\citet{ucan_bootstrap-based_2026} propose a bootstrap-stabilized k-means procedure that clusters features and selects representative features close to the cluster centers for subsequent fuzzy regression modeling. 
In contrast, our study clusters RF trees to select or construct representative trees, rather than combining forecasts through fuzzy functions or selecting representative features.

Taken together, existing approaches highlight that clustering and surrogate modeling are promising strategies for simplifying RFs.
Clustering has been used for data preprocessing and to explore heterogeneity among RF trees, without systematically informing the construction of surrogate models. 
Beyond RF surrogate modelling, related work uses feature selection to construct ensembles and fuzzy approaches for forecasting and feature selection. However, it has not addressed the interpretation of RFs via representative tree ensembles. 
Consequently, it remains unclear how clustering can be leveraged to build representative tree ensembles that balance predictive performance and model complexity, motivating a systematic investigation of such approaches.

\section{Methods}
\subsection{Ensembles of representative trees}\label{subsec:ensemblereptree}
One limitation of representing an RF using only a single tree is that it captures limited information, potentially overlooking important patterns and variable interactions across the forest. 
Using a small set of representative trees could provide a more comprehensive interpretation and improve prediction accuracy on test data. A naive approach would be to select the top $k$ MRTs with the highest average similarity to other trees in the RF. 
However, these trees may be similar to each other, potentially limiting the additional information gained from using multiple trees.	
An alternative is to partition the trees of the RF into $k$ clusters. 
With multiple clusters ($k>1$), one MRT or ART is obtained from each cluster to form an MRT ensemble (MRTE) or ART ensemble (ARTE), respectively. 
The special case $k=1$ recovers the original MRT and ART approaches, in which a single representative tree is obtained from the entire RF \citep{banerjee_identifying_2012,laabs_construction_2024}.
Ensemble predictions are obtained by aggregating the predictions of the individual trees the same way as in an RF, for example, by averaging them in regression.

\subsection{Simulation study design}
\textbf{Aim:} We investigate whether ARTEs represent the structure of RFs better than MRTEs. \\\\
\textbf{Data-generating mechanisms:}
We perform a simulation study based on an established approach \cite{seifert_surrogate_2019} resulting in a dataset for a regression task with a linear outcome. The outcome is simulated based on four effect variables using
\begin{align}
  y = X1 + X2 + X3 + X4 + N(0,0.2).
\end{align}
with $X1, \dots, X4 \sim N(0,1)$. We add noise variables $X5, \dots, X88 \sim N(0,1)$ and correlated variables for two effect variables ($X1$ and $X$2) and for two noise variables ($X5$ and $X6$). For each, three variables are simulated with correlations of $r$ = 0.9 to $X1$ and $X5$ and $r$ = 0.3 to $X2$ and $X6$. Each dataset contains a total number of $p$ = 100 variables for $n$ = 1000 and in total 200 datasets are simulated of which half are each utilized for training and testing purposes.\\\\
To additionally investigate nonlinear relationships and interaction effects, we performed a second simulation scenario following \citet{friedman_predictive_2008}. The outcome is generated as
\begin{align}
  y ={}& 9 \prod_{j=1}^{3} \exp\left[-3(1-X_j)^2\right]
         - 0.8 \exp\left[-2(X_4-X_5)\right] \notag\\
       & + 2\sin^2(\pi X_6) - 2.5(X_7-X_8) + \varepsilon.
\end{align}
The $p = 100$ independent variables are sampled uniformly from $\{0, 0.1, \ldots, 0.9\}$. The first eight variables are effect variables, while the remaining 92 are noise variables. Independent Gaussian noise $\varepsilon \sim N(0,\sigma^2)$ is added, with $\sigma^2$ set to half the empirical variance of the noise-free outcome values, yielding a signal-to-noise variance ratio of $2:1$. 
Each dataset contains 5000 training observations and 1000 independent test observations.\\\\
\textbf{Target:}
The targets of this study are the different ensembles and their predictions.\\\\
\textbf{Methods:}
Clustering-based ensembles of $k = 1, \dots 10$ MRTs or ARTs are generated by performing multidimensional scaling (MDS) on the tree distances, and applying $k$-means on the first two MDS components. 
From each identified cluster, an MRT can be selected or an ART can be created. We also include the naive approach of selecting the top $k = 1, \dots 10$ MRTs as a benchmark. As an additional baseline, we randomly select $k$ trees from the RF to form random tree (RT) ensembles.
All four methods are compared at the same prespecified $k = 1,\ldots,10$. ARTEs and MRTEs use identical cluster assignments, ensured by identical clustering inputs and a fixed random seed.
For ARTEs, we also include the option of automatically selecting the number of clusters ($k$) based on within-cluster variation, which must be greater than 10\% for an increase in $k$.
All representative trees are built using \emph{timbR} version 3.3.2 (\url{https://github.com/imbs-hl/timbR}) with the distance metric WSV \cite{laabs_identification_2024}. WSV is also used for clustering in the main analysis, following the findings of \citet{laabs_identification_2024}.
The required RF is grown with $num.trees = 500, mtry = \sqrt p = 10$, and $min.node.size = 100$ using \emph{ranger} version 0.16.0 \cite{wright_ranger_2017}.
For the ARTEs, the hyperparameters $min.bucket = 100$, and $probs\_quantiles = (0.25, 0.5, 0.75)$ of the ART algorithm are used to reduce runtime \cite{laabs_construction_2024}.%
The $imp.num.var$ parameter is not fixed, as the optimal number can vary between datasets. In addition, the optimal number is often unknown in subsequently used benchmark data. Instead, variable selection is performed using the \emph{Boruta} algorithm \cite{kursa_feature_2010}, following the recommendations of \citet{degenhardt_evaluation_2019}.

We conduct two sensitivity analyses for ARTEs on simulated data. In the first, we vary \emph{imp.num.var} in both simulation scenarios. In addition to using Boruta as described above, we consider a fixed number of 10 variables. A small fixed value may reduce computation time by avoiding a variable selection algorithm while still limiting the candidate splits. Here, 10 corresponds to the total number of effect and correlated variables in the linear scenario and thus represents an optimal choice if this information were known in advance. For comparison, we also set \emph{imp.num.var} to 100 allowing all available variables to be considered in both scenarios.

In the second, we investigate how grouping trees into clusters influences the performance of ARTEs. We compare WSV-based clustering from the main analysis with clustering based on the similarity of tree predictions on the training data and random assignment of the 500 RF trees to $k$ approximately equally sized clusters. 
One ART is constructed per cluster, using WSV for ART construction and Boruta variable selection in all three settings. 
Comparisons are performed at the same fixed ensemble sizes $k=1,\ldots,10$ in all scenarios.
At $k=1$, all three settings use the entire RF without clustering.
In each analysis, all other settings are kept at their main analysis values.
\\\\
\textbf{Performance measures:} Performance measures are reported across the 100 repetitions. All variables that are directly used to simulate the outcome are defined as effect variables, the ones that are correlated to them as correlated variables, and the remaining ones as noise variables.
The deviation from the RF is measured using the mean squared error (MSE) between the predictions of the MRTs and those of the RF on an independent test dataset. The deviation of the RF from MRTEs and ARTEs is calculated in the same way.
Overall accuracy is evaluated by the MSE between on the predictions and the true outcomes for each ensemble on an independent test dataset.
To jointly assess prediction accuracy and model complexity, we use a Bayesian information criterion (BIC)-type criterion adapted from \citet{grubinger_evtree_2014}, with their conservative default of $\alpha=1$ for the complexity penalty.
We extend their criterion to tree ensembles by using the total number of terminal nodes across all trees as a measure of ensemble complexity and evaluating the prediction error on independent test data. 
The criterion is calculated as
\begin{align}
  C_{\mathrm{BIC}}
  = n_{\mathrm{test}} \log(\mathrm{MSE}_{\mathrm{test}})
  + 4 \left(\sum_{j=1}^{k} M_j + 1\right)
    \log(n_{\mathrm{test}}),
  \label{eq:bic-type}
\end{align}
where $n_{\mathrm{test}}$ is the number of test observations, $\mathrm{MSE}_{\mathrm{test}}$ is the MSE of the ensemble predictions, and $M_j$ denotes the number of terminal nodes in tree $j$ of the ensemble.
Lower values indicate a more favorable trade-off between prediction accuracy and model complexity.
Similarity between the trees in the MRTEs and ARTEs is measured using SV and WSV.
Stability of individual predictions across the repetitions of the simulation is evaluated using an independently simulated common reference dataset of 1000 observations per scenario. For each method and ensemble-size setting, models trained in each of the 100 repetitions are applied to calculate predictions on this reference dataset. Individual level predictions are compared pairwise using mean squared differences and averaged across all pairs.  Lower values indicate greater prediction stability.
Relative frequency of how many of the effect, correlated, or noise variables are used as split variables is measured.

Computations used R 4.3.0 and \emph{batchtools} 0.9.17 \cite{lang_batchtools_2017} on a heterogeneous Linux cluster managed by Slurm. Cluster nodes differed in their hardware configurations, with 96 to 738 GB of RAM and 12 to
64 CPUs per node. One CPU was requested per job. All jobs were randomly distributed across the cluster nodes, with no preferential assignment of particular jobs to specific nodes.
We measured user CPU time for surrogate construction with \texttt{proc.time()}, excluding RF training and evaluation.
Because computations were performed on heterogeneous hardware, the reported runtimes indicate approximate computational effort and do not constitute a controlled comparison on identical processors.

\subsection{Benchmark experiment}
To evaluate performance under more realistic settings, we use 33 OpenML datasets from the AutoML Benchmark Regression Collection (ID 269) \cite{bischl_openml_2017,noauthor_openml_nodate}. These datasets contain between four and 4992 continuous and categorical variables. The number of observations ranges from 240 to 10 million (see \Cref{app:tab:bench:data}). 
Prior to model training, data filtering is performed in a procedure analogous to that described by \cite{laabs_construction_2024}. This involves removing outcome outliers that are more than five times the interquartile range from the median. Additionally, all observations with missing values in either the variables or the outcome are excluded. Variables with only one unique value are also removed from the datasets.
The RF model is trained using seven-fold cross-validation as in \cite{laabs_construction_2024}. Each RF comprises 500 trees. The number of candidate splitting variables uses the \emph{ranger} default, $mtry=\lfloor\sqrt{p}\rfloor$, where $p$ is the number of predictors after preprocessing, excluding the outcome. The minimum node size is set to $0.1\cdot n_{\mathrm{train}}$, where $n_{\mathrm{train}}$ is the number of training observations in the respective fold.
Only those datasets for which the out-of-bag $R^2$ exceeded 0.3 are retained for further analysis, as lower values indicate that the RF may not be a suitable model for the data, making its interpretation less meaningful.

ARTEs, MRTEs, MRTs, and RT ensembles are compared at the same fixed ensemble sizes $k=1,\ldots,10$. 
For the clustering-based ARTEs and MRTEs, $k$ is also the number of clusters. Automatic selection of $k$ is additionally evaluated for ARTEs using the relative within-cluster improvement criterion with a threshold of 10\% described above. Ensemble predictions are obtained as unweighted means of the individual tree predictions.
WSV is used both to calculate the distances between RF trees for clustering and to construct ARTs within the resulting clusters.
ART variable preselection uses Boruta ($imp.num.var=\text{automatic}$), as in the simulation.
The ART hyperparameters are $min.bucket=\lfloor0.1\cdot n_{\mathrm{train}}\rfloor$ and $probs\_quantiles=(0.25,0.5,0.75)$. 
Ensemble construction was subject to a time limit of 15 hours per repetition.
Methods are only included in the figures, if all repetitions for all $k$ finished in time.

As in the simulation study, we investigate how the cluster-generating mechanism influences ARTEs, comparing WSV-based clustering from the main analysis with clustering based on the similarity of tree-training predictions and random assignment to equally sized clusters. Stability analysis of individual predictions is also performed, similar to the simulation study. Prediction stability is calculated separately for each method. In the seven-fold cross-validation, each model was trained on six folds and tested on the remaining fold. Because the held-out folds differ between models, there is no common held-out dataset for comparing predictions across all seven models. We therefore obtained predictions from every model on the full preprocessed dataset. Using the full dataset was appropriate because the analysis focused on how much predictions varied across models for the same observations and not on the prediction accuracy. 
In each analysis, all other settings are kept at their main analysis values.

The evaluation is performed based on deviation from the RF, MSE on test data, the adapted BIC measure $C_{\mathrm{BIC}}$ and CPU time for each job. Model complexity is additionally described by the mean number of splits per tree and the total number of terminal nodes across the ensemble.
Prediction stability is assessed in the same way as in the simulations. Within each dataset, method, and ensemble-size setting, all fitted models predict the same complete preprocessed benchmark dataset.

\section{Results}
\subsection{Simulation study}

\begin{figure}[tpb]
  \centering 
  \includegraphics[width=\textwidth]{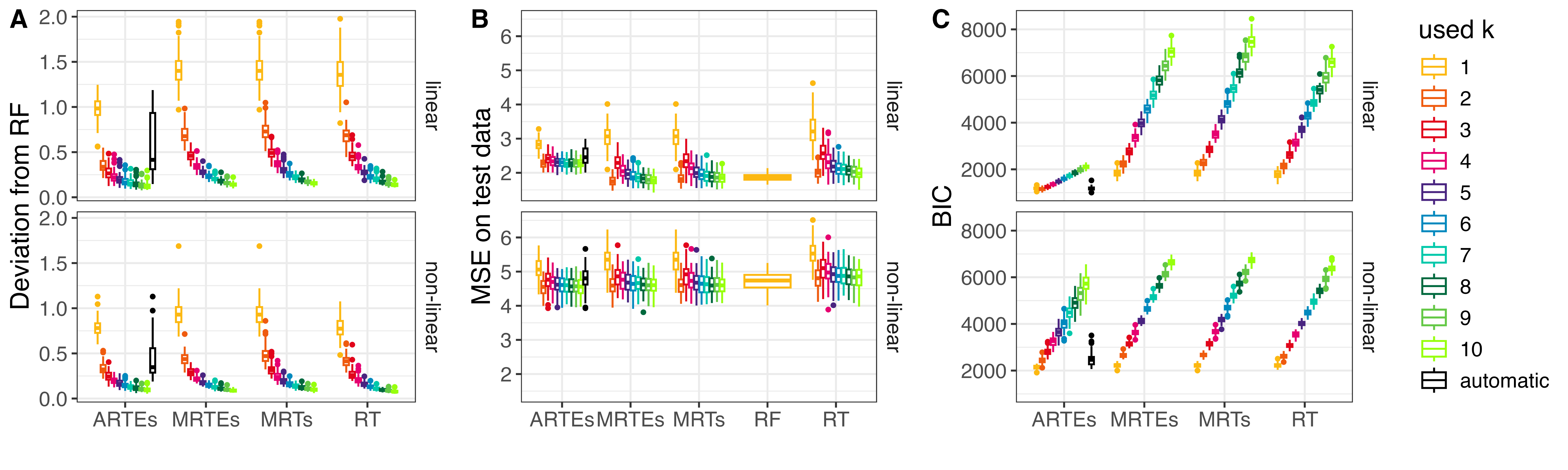}
  \caption{Comparison of the methods ARTE, MRTE, MRT, and RT ensembles over 100 repetitions in the linear (top) and nonlinear (bottom) simulation scenarios. Panels show the deviation from RF (A), the MSE on test data (B) and the BIC (C). The deviation from RF is measured as the MSE between the prediction of the RF and the respective models. Panel B additionally shows the MSE of the original RF. The BIC-type criterion combines test MSE with a penalty based on the total number of terminal nodes. Colors indicate fixed ensemble sizes $k=1,\ldots,10$, while white boxes denote automatic selection of $k$.}\label{fig:mse_ensembles}
\end{figure}

The deviation of the predictions from the RF decreases for increasing number of trees in the ensembles (\Cref{fig:mse_ensembles}A). ARTEs show lower deviation than MRTEs and MRTs at the same fixed $k$, particularly for small ensembles. Automatically selected ensembles show greater variability than ensembles with fixed $k$.

Higher numbers of $k$ also lead to smaller MSE and thus higher accuracy on independent data (\Cref{fig:mse_ensembles}B). In the linear scenario, ARTEs achieve lower MSE for single trees, whereas MRTEs and MRTs perform better for larger ensembles and approach the MSE of the RF. In the nonlinear scenario, differences between these three methods are smaller.

The BIC-type criterion generally increases with ensemble size in both simulation scenarios (\Cref{fig:mse_ensembles}C). In the linear scenario, ARTEs achieve substantially lower values than MRTEs and MRTs. In the nonlinear scenario, the differences are smaller, although ARTEs generally retain lower values at the same fixed $k$.

ARTs contain substantially fewer splits than the other trees in the linear scenario, whereas the differences are smaller in the nonlinear scenario (\Cref{app:fig:sim_num_splits}A). The total number of terminal nodes increases with $k$ for all methods (\Cref{app:fig:sim_num_splits}B).

\begin{figure}[tpb]
  \centering 
  \includegraphics[width=\textwidth]{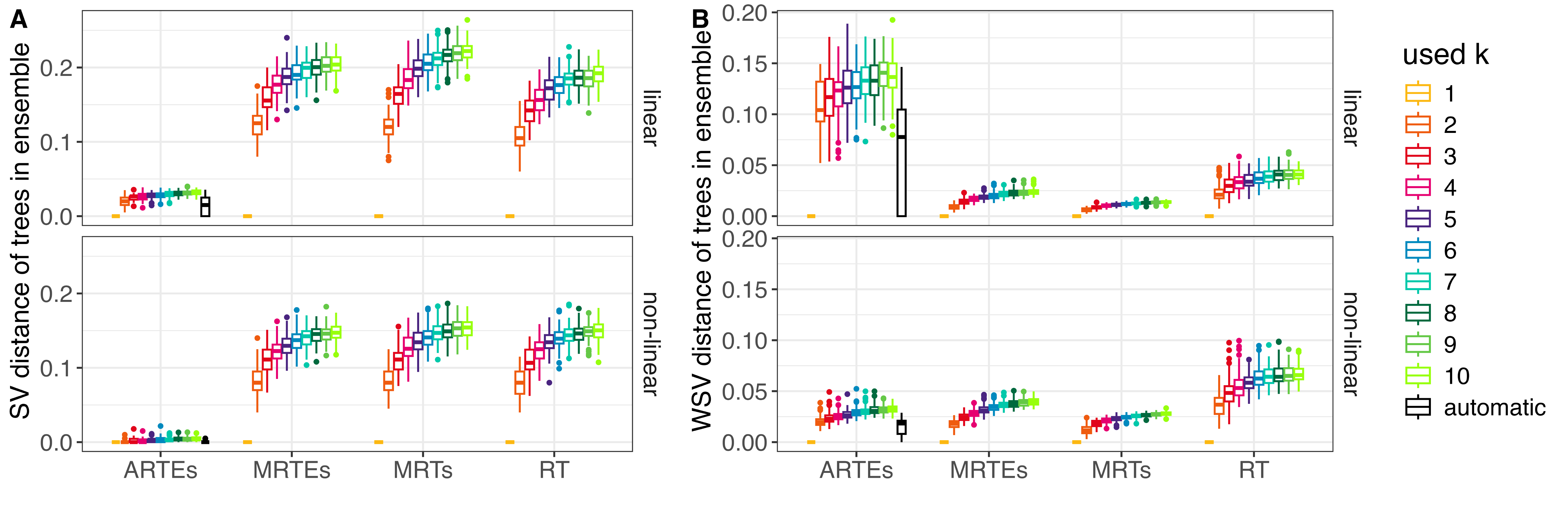}
  \caption{Mean pairwise distances between the trees of the ARTEs, MRTEs, MRTs, and RT ensembles over 100 repetitions in the linear (top) and nonlinear (bottom) simulation scenarios. The distances are measured using the metric splitting variables (SV, A) and weighted splitting variables (WSV, B). Colors indicate fixed ensemble sizes $k=1,\ldots,10$, while white boxes denote automatic selection of $k$.}\label{fig:dist_ensembles}
 \end{figure}

For both distance measures, the average distance between the trees increases with increasing $k$, i.e., with increasing number of trees (\Cref{fig:dist_ensembles}). 
In both simulation scenarios, SV distances are substantially lower for ARTEs than for MRTEs, MRTs, and RT ensembles (\Cref{fig:dist_ensembles}A), indicating greater similarity in the splitting variables used by ARTs.
The WSV distances show a scenario-dependent pattern (\Cref{fig:dist_ensembles}B). 
In the linear scenario, ARTEs exhibit the largest distances, whereas in the nonlinear scenario, their distances are lower than those of MRTEs and RT ensembles. Thus, similarity in variable usage does not necessarily imply similarity in the depth-weighted positions of these variables.

\begin{figure}[tpb]
  \centering 
  \includegraphics[width=\textwidth]{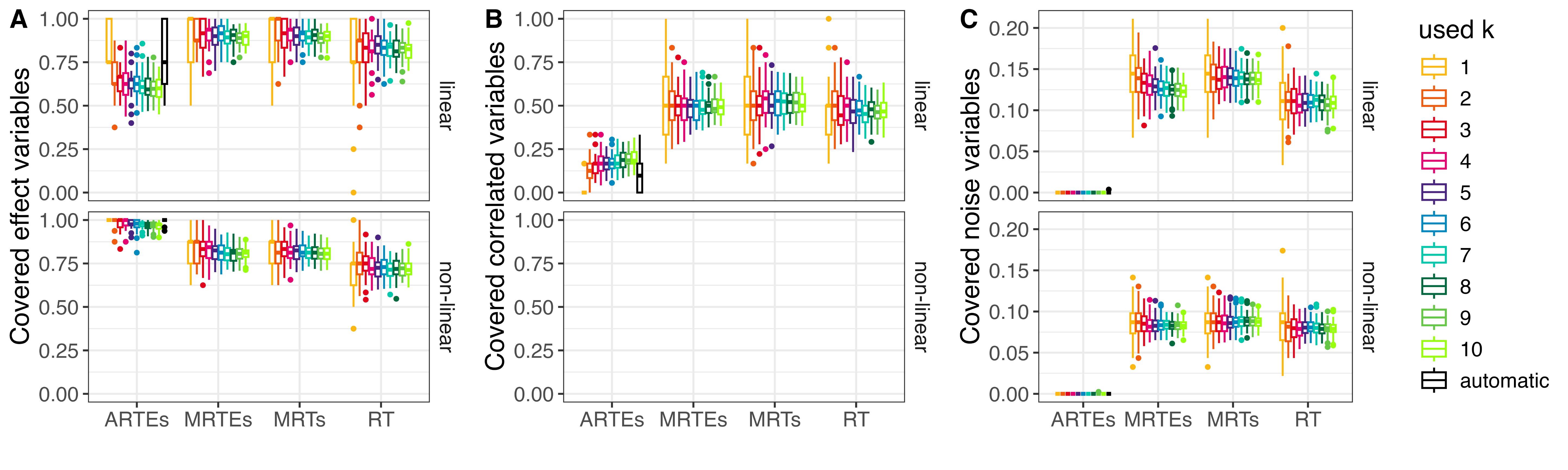}
  \caption{Relative frequency of effect (A), correlated (B), and noise (C) variables used as split variables in the trees of the ensembles of ARTEs, MRTEs, MRTs, and RT ensembles over 100 repetitions in the linear (top) and nonlinear (bottom) simulation scenarios. Effect variables are defined as all variables that are used to simulate the outcome, the correlated variables as the ones that are correlated with the effect variables and all others as noise variables. The panel for correlated variables is empty in the nonlinear scenario because no correlated variables were generated. Colors indicate fixed ensemble sizes $k=1,\ldots,10$, while white boxes denote automatic selection of $k$.}\label{fig:covered_vars_ensembles}
\end{figure}

Variable usage differs between the simulation scenarios (\Cref{fig:covered_vars_ensembles}). In the linear scenario, ARTEs use fewer effect and correlated variables than MRTs and MRTEs, but they are less deep, which inherently limits the number of variables they can include (\Cref{fig:covered_vars_ensembles}).  
In the nonlinear scenario, ARTs cover a larger proportion of effect variables. In both scenarios, ARTs rarely use noise variables, whereas the other methods show substantially greater noise variable usage.

Prediction stability improves with increasing ensemble size in both simulation scenarios, as indicated by decreasing mean pairwise prediction MSE (\Cref{app:fig:sim_prediction_stability}). 
At the same fixed $k$, ARTEs generally show smaller prediction differences than MRTEs and MRTs, particularly for small ensembles. 
ARTEs with automatically selected $k$ show greater variability than those with fixed ensemble sizes.

The median CPU time is 11.74 minutes for ARTEs with Boruta variable selection, compared with 0.0100 minutes for MRTEs and 0.0083 minutes for MRTs. For RT ensembles, the CPU time was so short that the timer recorded it as zero in at least half of the runs.

Varying \emph{imp.num.var} produces similar deviations from the RF, test MSE values, and prediction distances at the same fixed ensemble size (\Cref{app:fig:sim_imp_num_var}). In the nonlinear scenario, allowing all 100 variables increases noise variable usage and SV distances and generally leads to larger ensembles in terms of terminal nodes and higher BIC-type values. Boruta requires substantially more CPU time than either fixed setting.

Using prediction-based rather than WSV-based clustering yields broadly similar results in both simulation scenarios (\Cref{app:fig:sim_cluster_metric}). Prediction-based clustering generally produces slightly lower deviations from the RF and test MSE values, while variable usage and model complexity remain similar.
Random clustering generally produces greater deviations from the RF and larger prediction differences across repetitions, indicating lower fidelity and prediction stability.
Differences in test MSE are comparatively small. 
In the linear scenario, random clustering yields similar ensemble complexity but lower pairwise SV and WSV distances, greater coverage of effect variables, and lower coverage of correlated variables.
In the nonlinear scenario, random clustering generally produces more terminal nodes and higher BIC-type values at larger $k$.
Noise variable usage remains low under all three clustering procedures, and runtime distributions largely overlap.

\subsection{Benchmark experiment}
For five datasets, at least one ART ensemble run exceeded the 15-hour time limit. The figures therefore omit all ART results for those datasets but retain the results for the other methods.
Missing ART results by dataset and ensemble size are summarized in \Cref{app:tab:benchmark_missing_arte}.
Numerical benchmark results for all 19 included datasets are provided in the Supplementary Data (CSV files). These report fidelity to the RF, test MSE, total terminal-node counts, and user CPU time by dataset, method, and ensemble size.
The combined file additionally reports the number of available cross-validation folds, allowing incomplete results to be identified.

\begin{figure}[tbp]
  \centering 
  \includegraphics[width=\textwidth]{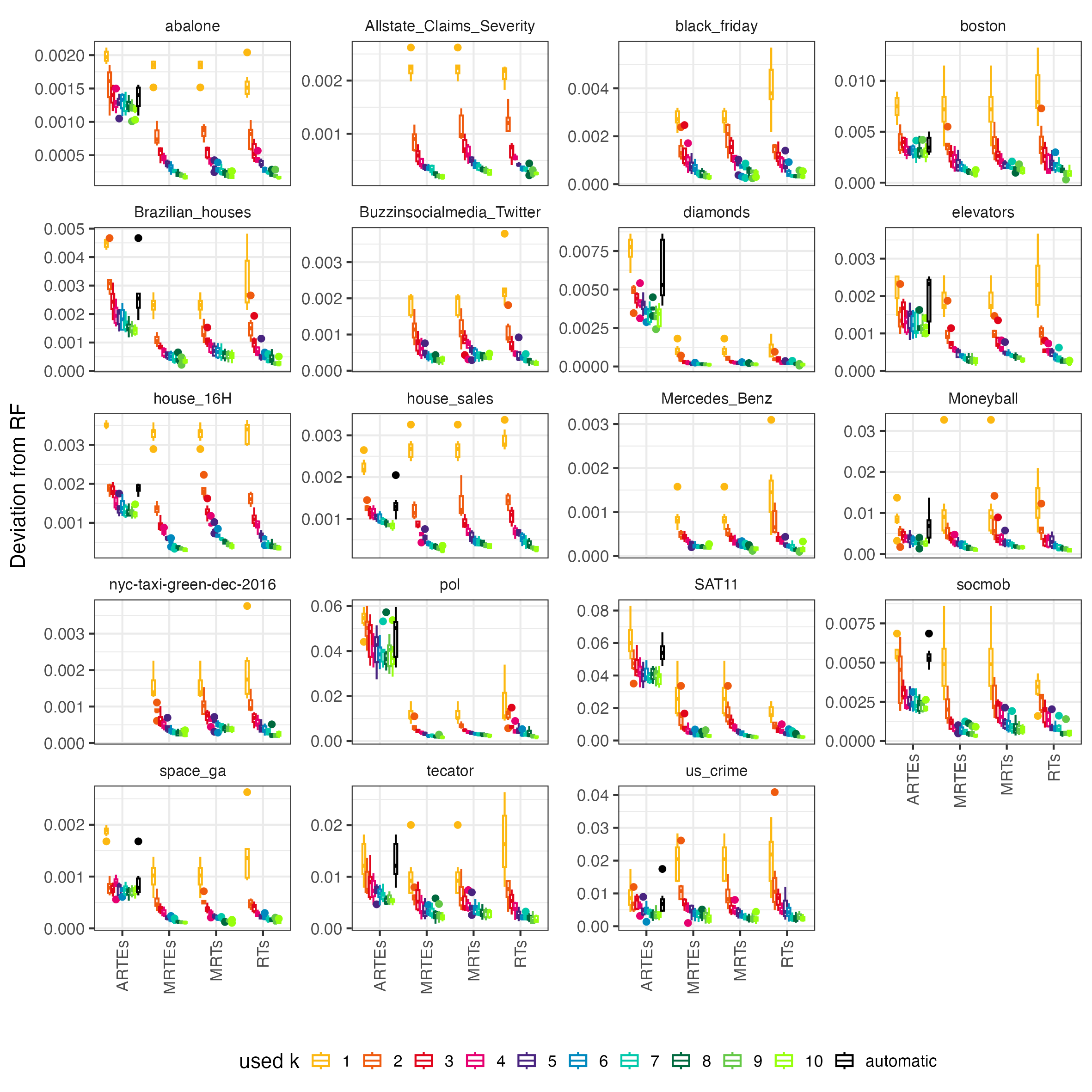}
  \caption{Deviation from RF predictions for ARTEs, MRTEs, MRTs, and RT ensembles in seven-fold cross-validation. The deviation from RF is measured as the MSE between the prediction of the RF and the respective models. Each panel represents a benchmark dataset. Colors indicate fixed ensemble sizes $k=1,\ldots,10$, while white boxes denote automatic selection of $k$.}\label{fig:bench_dev_ensembles}
\end{figure}

For most datasets, MRTEs show the least deviation from the RF, followed closely by MRTs, as shown in \Cref{fig:bench_dev_ensembles}.
ARTEs generally show greater deviation at larger fixed ensemble sizes, while differences between methods are smaller for single-tree models. RT ensembles show broadly similar trends to MRTs and MRTEs, although their deviation is often greater for small ensembles.

For most datasets, both MRTs and MRTEs show lower MSE values than ARTEs, which approaches the MSE of the RF as $k$ increases, as shown in \Cref{fig:bench_mse_ensembles}. 
In general, the MSE decreases with increasing values of $k$, consistent with the results observed in the simulation study. RT ensembles generally achieve similar MSE values to MRTs and MRTEs at larger $k$.

\begin{figure}[tbp]
  \centering 
  \includegraphics[width=\textwidth]{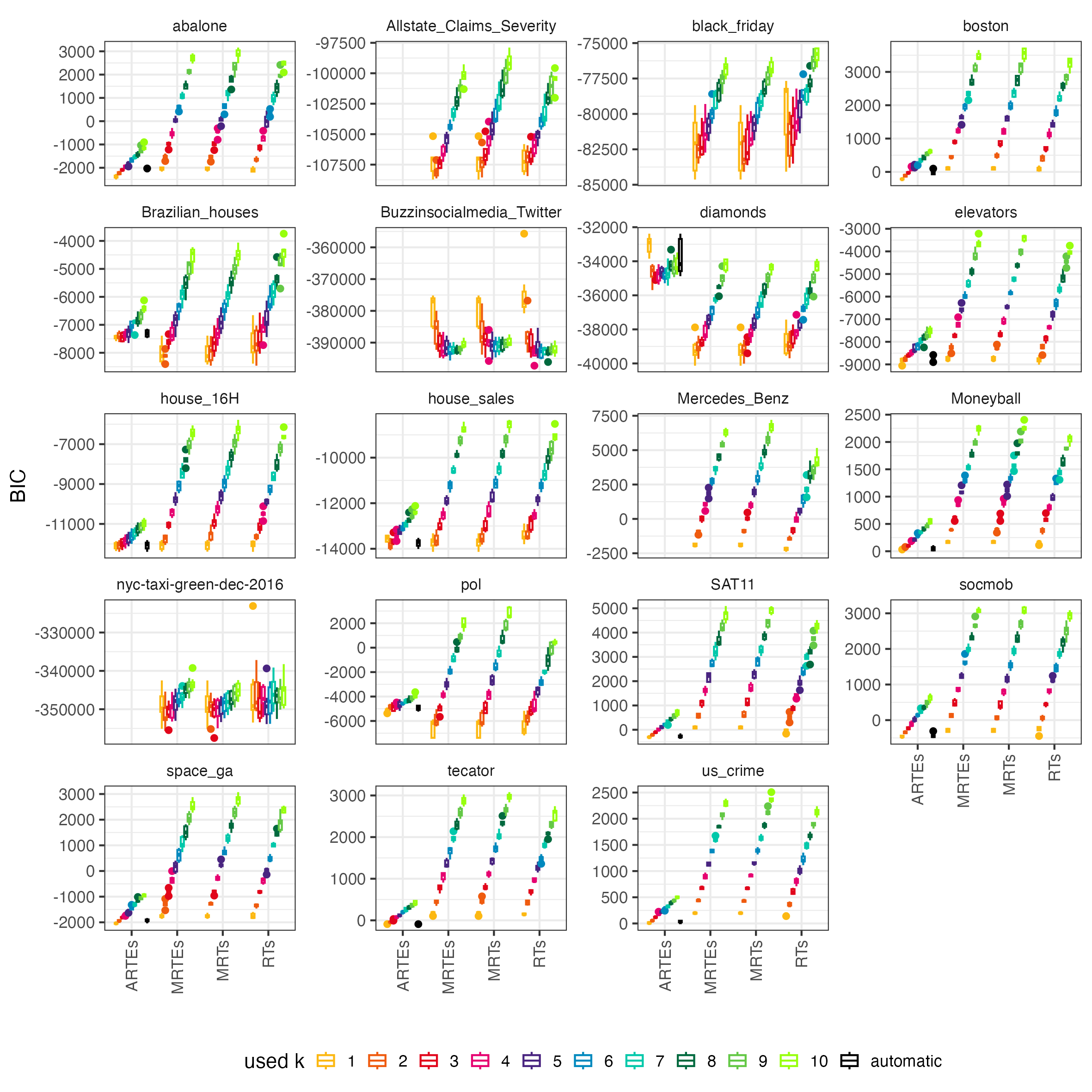}
  \caption{BIC-type criterion $C_{\mathrm{BIC}}$ for ARTEs, MRTEs, MRTs, and RT ensembles in seven-fold cross-validation. The criterion combines MSE on the held-out fold with a penalty based on the total number of terminal nodes. Each panel represents a benchmark dataset. Colors indicate fixed ensemble sizes $k=1,\ldots,10$, while white boxes denote automatic selection of $k$.}\label{fig:bench_bic_ensembles}
\end{figure}

ARTEs generally achieve lower BIC-type values than MRTs and MRTEs at larger fixed ensemble sizes, as shown in \Cref{fig:bench_bic_ensembles}, reflecting their lower total number of terminal nodes.
For most datasets, the BIC-type criterion increases with $k$ as ensemble complexity grows, although this pattern is less pronounced for some datasets.
Further details on the number of splits and terminal nodes can be found in the appendix (\Cref{app:fig:bench_num_splits}).

Prediction stability generally improves with increasing ensemble size, as indicated by decreasing prediction distances across cross-validation folds (\Cref{app:fig:bench_prediction_stability}).
ARTEs often show smaller prediction distances for small fixed ensembles, whereas differences between methods become smaller at larger $k$. Automatically selected ARTEs show greater prediction distances than larger fixed ART ensembles for several datasets.

The runtime of the RT ensembles (below 0.00 minutes), MRTs (median: 0.01 minutes) and MRTEs (median: 0.01 minutes) is shorter than that of the ARTEs with variable selection (median: 12.05 minutes).

WSV-based and prediction-based clustering often yield lower deviations from the RF than random tree clusters (\Cref{app:fig:bench_cluster_fidelity})
This difference is particularly pronounced for boston, house\_16H, socmob, and space\_ga, whereas differences are smaller at larger $k$ for datasets such as diamonds and Moneyball.
Neither WSV-based nor prediction-based clustering consistently outperforms the other across datasets.
Differences in test MSE are generally less pronounced and depend on the dataset and ensemble size (\Cref{app:fig:bench_cluster_accuracy}).
Prediction stability improves with increasing ensemble size under all three clustering procedures, but neither WSV-based nor prediction-based clustering consistently yields smaller prediction distances than random clustering (\Cref{app:fig:bench_cluster_stability}).
Additional results on ensemble complexity, the BIC-type criterion,
within-ensemble SV and WSV distances, and runtime are provided in
\Cref{app:fig:bench_cluster_complexity,app:fig:bench_cluster_bic,app:fig:bench_cluster_sv,app:fig:bench_cluster_wsv,app:fig:bench_cluster_runtime}.

\section{Discussion}
The performance of MRTEs and MRTs is very similar in terms of deviation from the RF, MSE or BIC-type criterion on test data, and variable usage, with MRTEs showing slightly better performance and using fewer noise variables.
RT ensembles often approach the performance of MRTs and MRTEs at larger $k$, suggesting that averaging multiple trees contributes to the observed improvements.
For ARTEs, the comparison with random tree partitions indicates that similarity-based clustering contributes to fidelity beyond the gains obtained by constructing and averaging multiple ARTs.
However, its advantages for prediction accuracy and prediction stability are less consistent across benchmark datasets.

In contrast, the ARTEs achieve lower BIC-type values due to their smaller trees, resulting in more compact  and thus interpretable models.
In the linear simulation, ARTEs use smaller proportions of effect and correlated variables, while they cover more effect variables in the nonlinear scenario. In both scenarios, they rely far less on noise variables, which reduces the risk of misleading conclusions regarding variable importance.
At the same time, MRTEs tend to achieve higher predictive accuracy, highlighting a trade-off between accuracy and model complexity. This trade-off is particularly evident in the benchmark experiment: at the same larger fixed $k$, ARTEs generally use fewer terminal nodes and achieve lower BIC-type values, but show greater deviation from the RF and higher test MSE than MRTs and MRTEs (\Cref{fig:bench_dev_ensembles,fig:bench_bic_ensembles}, \Cref{fig:bench_mse_ensembles,app:fig:bench_num_splits}).
ARTEs therefore offer benefits when compact surrogate models are a priority and the observed loss of fidelity and prediction accuracy is acceptable for the application. Lower BIC-type values alone do not establish that fidelity is sufficient. Deviation from the RF and test MSE should be considered alongside model complexity.

However, the information gained from using ARTEs may decrease as $k$ increases, given that the ARTE trees are more similar to each other than MRTE trees. Therefore, increasing the number of trees does not necessarily lead to additional structural insight, and excessively large ensembles reduce comprehensibility.
This supports the use of small ART ensembles when the aim is a compact representation with a limited number of decision paths, provided that their fidelity and prediction accuracy meet the requirements of the application.
In the simulations, the greater variability of the MRTE split variables is likely due to the use of noise variables in the MRTs, which increases the diversity of the MRTE but is not beneficial for interpretation.

Prediction stability generally improves with ensemble size, and ARTEs often show smaller prediction differences for small ensembles. This indicates more reproducible predictions, which provide a useful basis for interpretation, although the stability of tree structures requires separate assessment.

The sensitivity analysis indicates that the choice of \emph{imp.num.var} affects noise variable usage and computational effort more than prediction performance in the investigated scenarios. 
Although Boruta increases runtime, consistent with the findings of \citet{degenhardt_evaluation_2019}, it avoids fixing the number of candidate variables using knowledge available only in simulation settings.

The comparison with random tree clustering helps distinguish the contribution of similarity-based clustering from that of constructing and averaging multiple ARTs.
In the simulations, similarity-based clustering generally improves fidelity and prediction stability.
On benchmark datasets, fidelity often improves, whereas advantages in prediction accuracy and prediction stability are less consistent.
Neither WSV-based nor prediction-based clustering consistently outperforms the other across benchmark datasets.

Several limitations should be considered when interpreting these findings. 
The reduced use of noise variables and smaller tree sizes should be interpreted in the context of the different construction procedures. ARTEs use variable preselection and split constraints, whereas MRTs and MRTEs select existing RF trees without equivalent restrictions. These components were investigated in the original ART study \cite{laabs_construction_2024}. Although our sensitivity analysis varies variable preselection, it does not fully separate its contribution from those of split constraints and ART construction.

The comparison with random tree clustering assesses the contribution of similarity-based clustering at matched ensemble sizes in both simulations and benchmark experiments.
However, random clusters are approximately equal in size, whereas similarity-based clusters may differ in size.
The comparison therefore does not fully separate the effects of similarity-based cluster composition from those of cluster size.

Our stability analyses are limited to variability in individual predictions. Further work is needed to assess the stability of cluster assignments and representative tree structures, as trees can yield similar predictions through different decision paths.

The present findings are limited to the investigated simulation settings and benchmark datasets. Filtering datasets by RF OOB $R^2$ may introduce selection bias, but interpreting predictive patterns is meaningful only when the RF captures a signal. On larger datasets, the time-consuming Boruta variable selection prevented completion of some analyses using ARTEs within the 15-hour job limit. ARTEs results were incomplete for five benchmark datasets (tab. \ref{app:tab:benchmark_missing_arte}), and only complete ARTEs results are shown. Comparisons involving ARTEs may therefore be biased toward computationally manageable datasets, limiting generalizability.

These limitations suggest several directions for future work. 
Matching random cluster sizes to those of the similarity-based clusters would help isolate the contribution of cluster composition in future comparisons.
In addition, user studies are needed to determine whether these ensembles improve understanding of RF predictions, and independent reference data would help assess prediction stability and prediction-based clustering. 
Although the simulations cover both linear and nonlinear outcome relationships, they do not systematically vary the proportion of noise variables or explore more complex correlation structures among predictors. Investigating these aspects would be an interesting direction for future work and would help assess the generalizability of the findings.
Vita could provide a faster alternative for high-dimensional datasets \cite{degenhardt_evaluation_2019}, but its effects on variable usage, fidelity, prediction accuracy, and completion rates in ART ensembles require evaluation.

Feature-space manipulation, as explored in recent ensemble methods \cite{jiang_neighborhood_2025,jiang_ensemble_2026}, could offer an additional mechanism of promoting diversity among ARTs. 
Future work could also investigate cluster-specific feature subsets for ART construction and assess whether they improve the trade-off between fidelity to the RF, prediction accuracy, and model complexity. 
The bootstrap-stabilized feature selection approach of \citet{ucan_bootstrap-based_2026} provides a methodological starting point for investigating whether repeated clustering and aggregation can improve the robustness of representative tree ensembles. 

In terms of clustering, other methods could be explored to create ART ensembles through optimization, with the objective of constructing $k$ regular ARTs that are as similar as possible to the entire RF, while ensuring that they are as diverse as possible.
This approach would balance similarity to the RF and sufficient diversity among the trees, which could improve both MSE and deviation from the RF.
Such extensions may further improve the balance between predictive accuracy and structural representation. This is particularly relevant because previous work has mainly used clustering for exploratory purposes rather than for systematically constructing surrogate models.

Our study characterizes the trade-offs of representative tree ensembles as RF surrogates and assesses the contribution of similarity-based tree clustering.
Compared with random tree clustering, similarity-based clustering often improves ART ensemble fidelity, whereas its benefits for prediction accuracy and prediction stability are less consistent across benchmark datasets.
ART ensembles provide compact surrogates when reduced complexity is a priority and the resulting fidelity and prediction accuracy are sufficient for the intended application.
Increasing ensemble size can improve prediction performance and stability, but also increases complexity and may make interpretation more difficult.

\backmatter





\bmhead{Acknowledgments}
This study was funded by the Medical Section of the University of Luebeck (J01-2024 to BL).

\section*{Declarations}
\bmhead*{Funding:} This study was funded by the Medical Section of the University of Luebeck (J01-2024 to BL).\\
This work is partially supported by the Ministry of Science and Culture of Lower Saxony through funds from the program zukunft.niedersachsen of the Volkswagen Foundation for the 'CAIMed – Lower Saxony Center for Artiﬁcial Intelligence and Causal Methods in Medicine' project (grant no. ZN4257). 

\bmhead*{Conflict of interest/Competing interests:} None 
\bmhead*{Ethics approval and consent to participate:} Not applicable, no respondents involved. 
\bmhead*{Consent for publication:} All authors consent to publication.
\bmhead*{Data availability:} The simulated data used for the publication is available on GitHub (\url{https://github.com/imbs-hl/Cluster_ART_paper}). The benchmark data is freely available at OpenML \cite{noauthor_openml_nodate} with ID 269 \url{https://openml.org/search?type=benchmark&sort=tasks_included&
study_type=task&id=269}.
\bmhead*{Materials availability:} The code to reproduce the study is available on GitHub (\url{https://github.com/imbs-hl/Cluster_ART_paper}). 
Additional figures and tables supporting the simulation study and benchmark experiment are provided in the Supplementary Information (Figures A1–A4 and B5–B15; Tables B1–B2).
Supplementary Data comprise numerical benchmark results in CSV format, provided as a combined file and separate files for each fixed ensemble size and automatic ensemble-size selection.
\bmhead*{Code availability:} The code to reproduce the study and the simulated data is available on GitHub (\url{https://github.com/imbs-hl/Cluster_ART_paper}).
\bmhead*{Author contribution:} L.L.K.: conceptualization, data curation, formal analysis, investigation, methodology, software, visualisation, writing - original draft, writing - review \& editing; M.W.: investigation, writing - review \& editing; S.S.: conceptualization, supervision, writing - review \& editing; B.H.L.: conceptualization, funding acquisition, project administration, supervision, writing - review \& editing







\FloatBarrier
\begin{appendices}

\section*{Overview of supplementary material}

\begingroup
\setlength{\parindent}{0pt}
\setlength{\parskip}{0.4em}

\newcommand{\suppentry}[3]{%
  \par\noindent
  \hangindent=1.5em
  \hangafter=1
  #1~\ref{#2}: #3\nobreak\dotfill
  \mbox{\pageref{#2}}\par
}

\subsection*{Supplementary Figures}

\subsubsection*{Simulation study}

\suppentry{Figure}{app:fig:sim_num_splits}
  {Mean number of splits and total number of terminal nodes}

\suppentry{Figure}{app:fig:sim_prediction_stability}
  {Prediction stability across simulation repetitions}

\suppentry{Figure}{app:fig:sim_imp_num_var}
  {Sensitivity analysis of \emph{imp.num.var}}

\suppentry{Figure}{app:fig:sim_cluster_metric}
  {Sensitivity analysis of clustering methods}

\subsubsection*{Benchmark experiment}

\suppentry{Figure}{fig:bench_mse_ensembles}
  {Predictive accuracy on held-out folds}

\suppentry{Figure}{app:fig:bench_num_splits}
  {Mean number of splits and total number of terminal nodes}

\suppentry{Figure}{app:fig:bench_prediction_stability}
  {Prediction stability across cross-validation folds}

\suppentry{Figure}{app:fig:bench_cluster_fidelity}
  {Comparison of clustering methods: fidelity to RF predictions}

\suppentry{Figure}{app:fig:bench_cluster_accuracy}
  {Comparison of clustering methods: predictive accuracy}

\suppentry{Figure}{app:fig:bench_cluster_stability}
  {Comparison of clustering methods: prediction stability}

\suppentry{Figure}{app:fig:bench_cluster_complexity}
  {Comparison of clustering methods: total number of terminal nodes}

\suppentry{Figure}{app:fig:bench_cluster_bic}
  {Comparison of clustering methods: BIC}

\suppentry{Figure}{app:fig:bench_cluster_sv}
  {Comparison of clustering methods: splitting-variable distances}

\suppentry{Figure}{app:fig:bench_cluster_wsv}
  {Comparison of clustering methods: weighted splitting-variable distances}

\suppentry{Figure}{app:fig:bench_cluster_runtime}
  {Comparison of clustering methods: runtime}

\subsection*{Supplementary Tables}

\suppentry{Table}{app:tab:benchmark_missing_arte}
  {Missing ART ensemble results in the benchmark experiment}

\suppentry{Table}{app:tab:bench:data}
  {Overview of benchmark datasets and inclusion criteria}

\endgroup

\clearpage

\section{Additional figures of simulation study}
\label{sec:appendix}

\begin{figure}[ht!]
  \centering 
  \includegraphics[width=\textwidth]{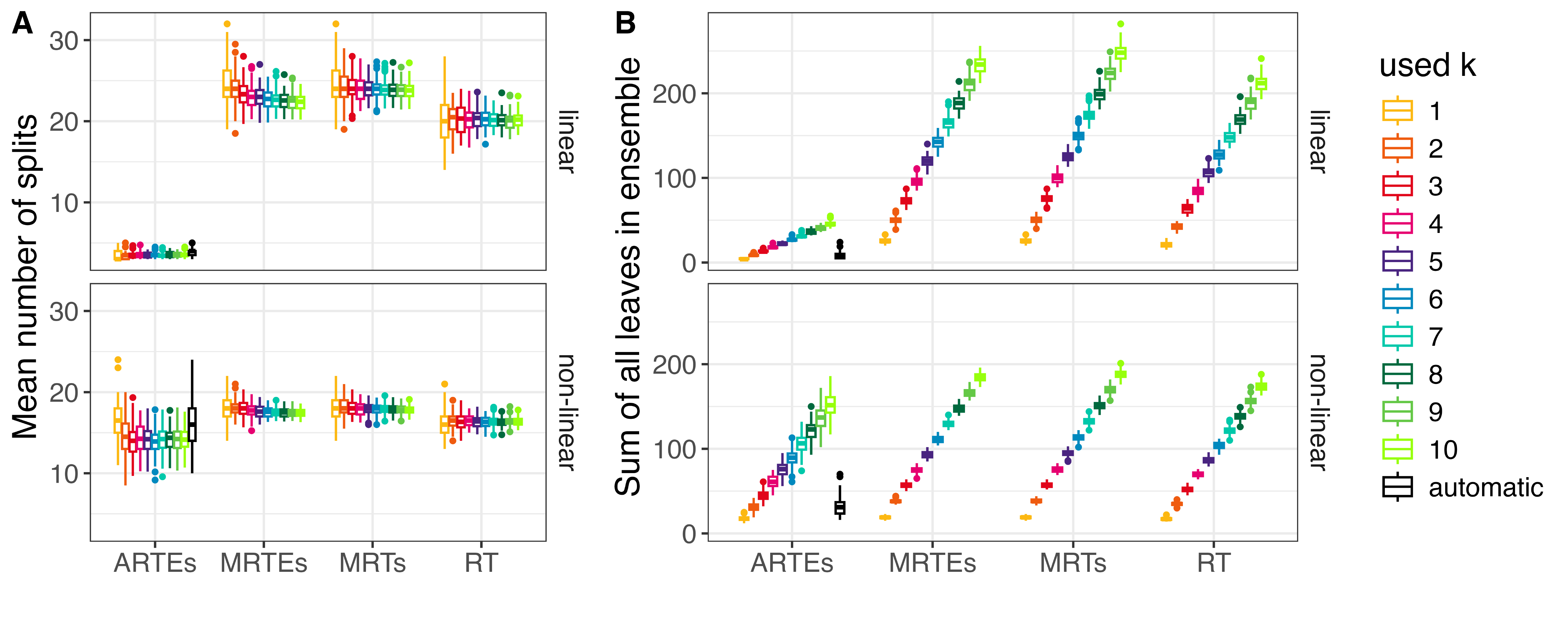}
  \caption{Mean number of splits per tree (A) and total number of terminal nodes across the ensemble (B) for ARTEs, MRTEs, MRTs, and RT ensembles over 100 repetitions in the linear (top) and nonlinear (bottom) simulation scenarios. Colors indicate fixed ensemble sizes $k=1,\ldots,10$, while white boxes denote automatic selection of $k$.}\label{app:fig:sim_num_splits}
 \end{figure}

\begin{figure}[p]
  \centering
  \includegraphics[width=\textwidth]{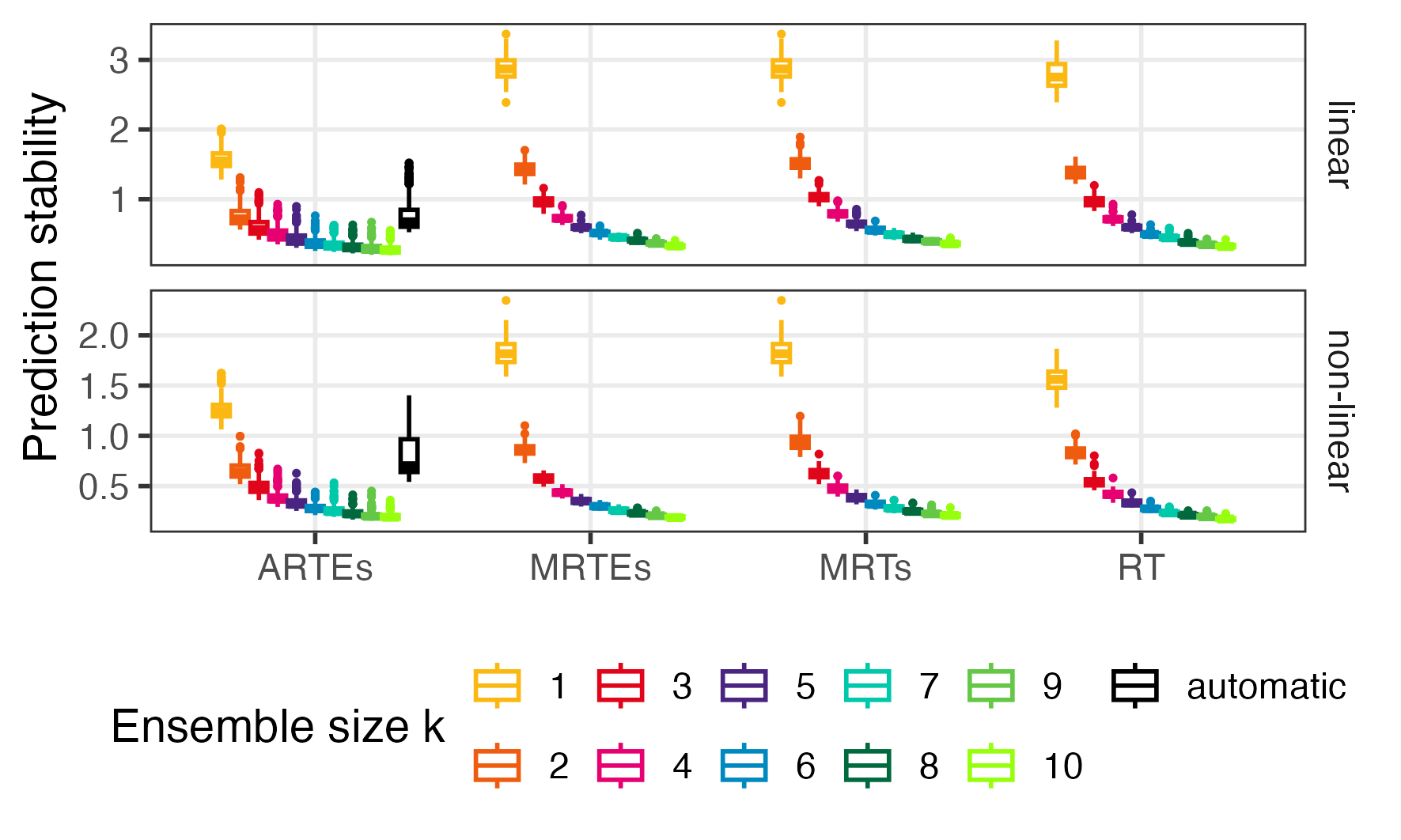}
  \caption{Prediction stability over 100 simulation repetitions in the linear (top) and nonlinear (bottom) scenarios. Each boxplot shows the distribution of model-specific mean pairwise prediction MSE values calculated on a common, newly simulated reference dataset of 1000 observations per scenario. Lower values indicate greater prediction stability. Colors indicate fixed ensemble sizes $k=1,\ldots,10$, while white boxes denote automatic selection of $k$}
  \label{app:fig:sim_prediction_stability}
\end{figure}

\begin{figure}[p]
  \centering
  \includegraphics[width=\linewidth,height=0.85\textheight,keepaspectratio]{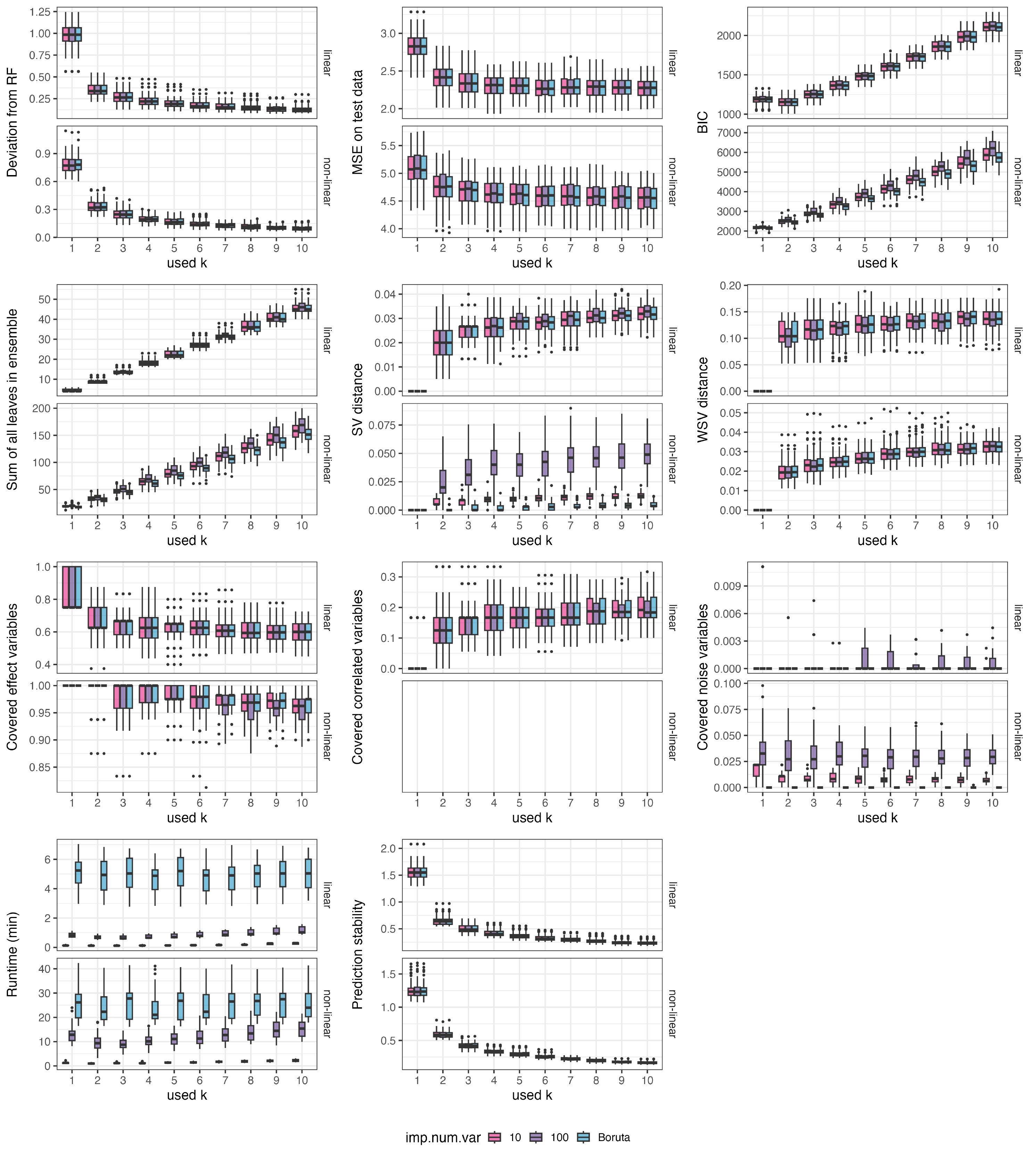}
  \caption{Sensitivity analysis of \emph{imp.num.var} for ARTEs over 100 repetitions. Boxplots compare settings of 10, 100, and automatic variable selection using Boruta at fixed ensemble sizes $k=1,\ldots,10$. Each metric is shown separately for the linear and nonlinear simulation scenarios. The correlated variable panel is empty for the nonlinear scenario because no correlated variables were generated. SV and WSV distances are shown only for ensembles with at least two trees. Prediction distance measures stability across simulation repetitions using the same reference dataset of 1000 observations per scenario for all three settings. Lower prediction distances indicate greater stability. Runtime is user CPU time in minutes.}
  \label{app:fig:sim_imp_num_var}
\end{figure}

\begin{figure}[p]
  \centering
  \includegraphics[width=\linewidth,height=0.78\textheight,keepaspectratio]{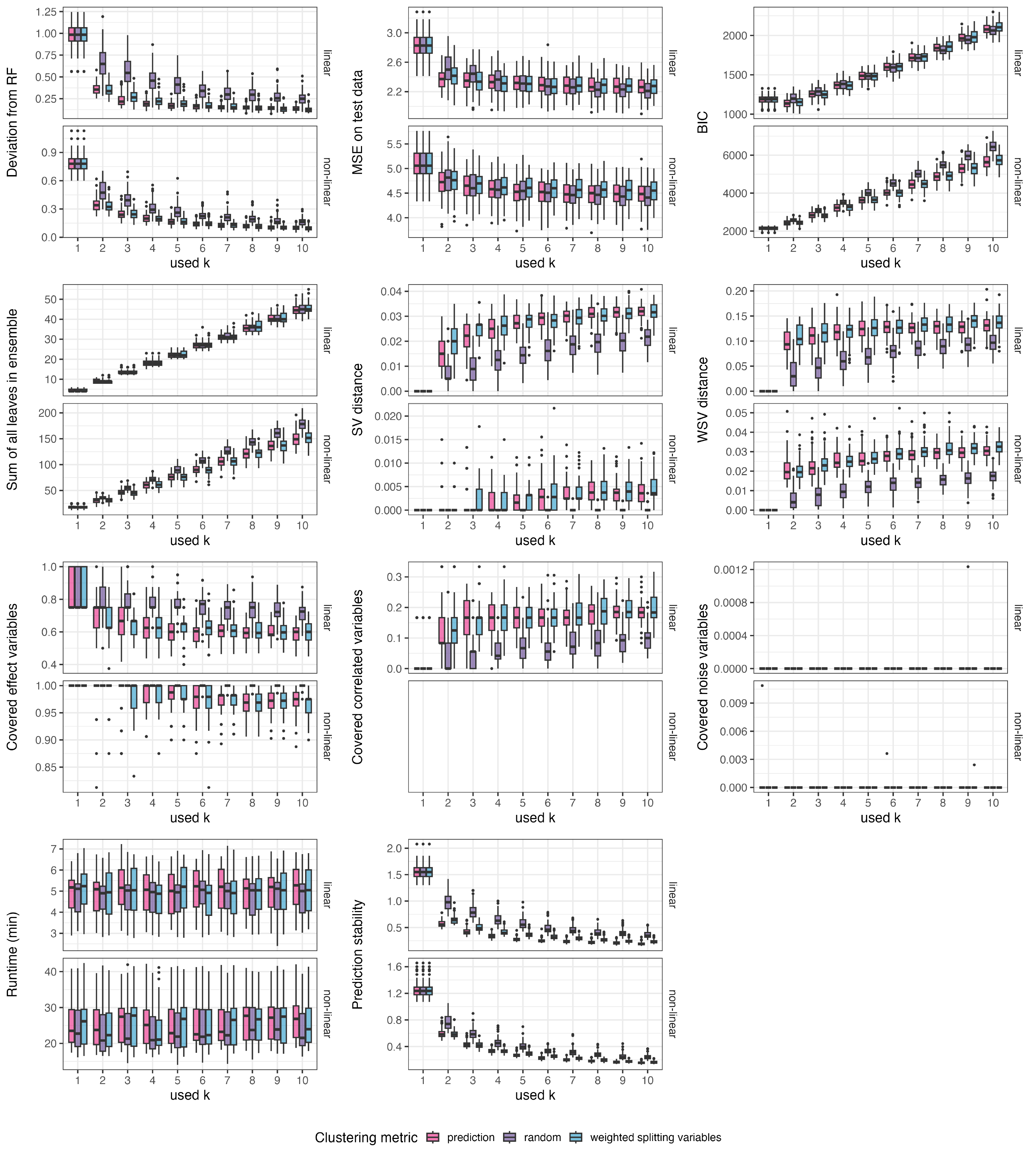}
  \caption{Sensitivity analysis comparing prediction-based clustering (yellow), random clustering (pink), and WSV-based clustering (purple) for ARTEs over 100 repetitions at fixed ensemble sizes $k=1,\ldots,10$. Each measure is shown separately for the linear and nonlinear simulation scenarios. Prediction-based clustering uses predictions of the RF trees on the training observations. Random clustering assigns RF trees to approximately equally sized randomly drawn clusters. ART construction uses WSV and Boruta variable selection in all three settings. Performance is evaluated on independent test data. SV and WSV distances describe mean pairwise distances within each ensemble and are shown for $k\geq2$. Variable coverage is the proportion of variables of each type used for splitting. No correlated variables were generated in the nonlinear scenario. Prediction stability is measured by mean pairwise prediction MSE across repetitions on a common, newly simulated reference dataset of 1000 observations per scenario. Lower values indicate greater stability.}
  \label{app:fig:sim_cluster_metric}
\end{figure}

\FloatBarrier
\section{Additional figures and tables of benchmark experiment}

\begin{figure}[ht!]
   \centering 
   \includegraphics[width=\textwidth]{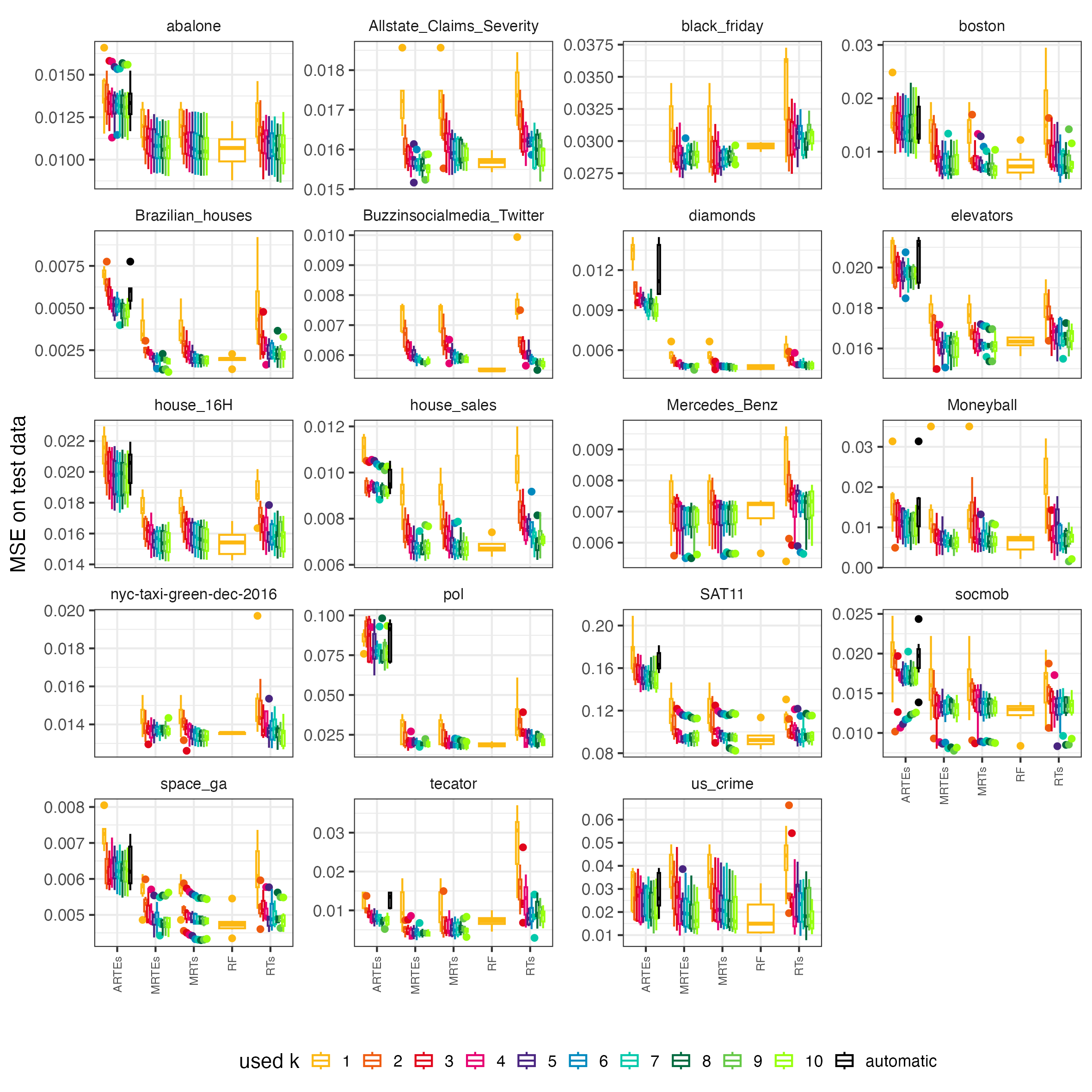}
   \caption{MSE on the held-out folds for ARTEs, MRTEs, MRTs, and RT ensembles in seven-fold cross-validation. The MSE of the original RF is shown for comparison. Each boxplot summarizes results across the seven cross-validation folds, and each panel represents a benchmark dataset. Colors indicate fixed ensemble sizes $k=1,\ldots,10$, while white boxes denote automatic selection of $k$.}\label{fig:bench_mse_ensembles}
 \end{figure}

\begin{figure}[ht!]
  \centering 
  \includegraphics[width=\textwidth]{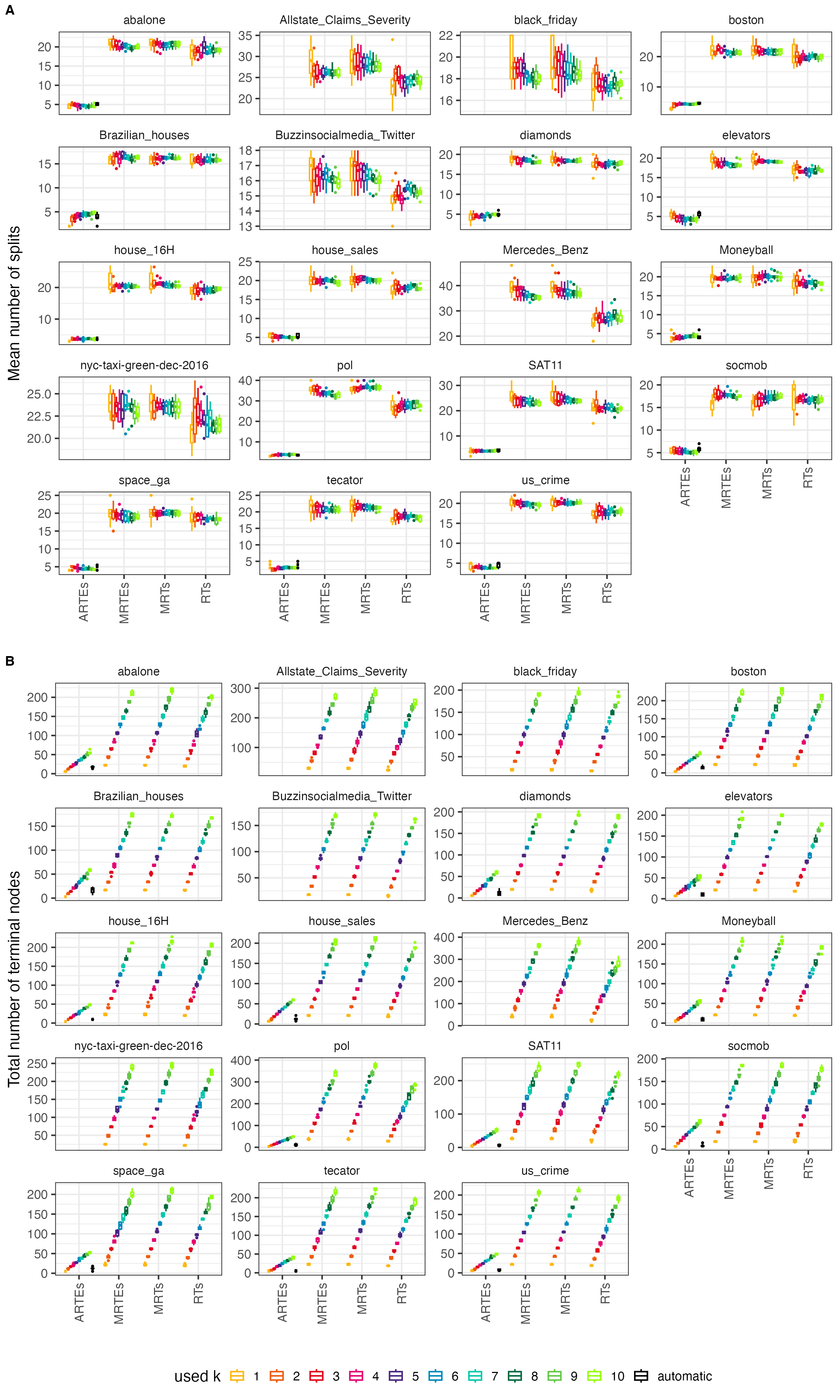}
  \caption{Mean number of splits per tree (A) and total number of terminal nodes across the ensemble (B) for ARTEs, MRTEs, MRTs, and RT ensembles in seven-fold cross-validation. Each boxplot summarizes results across the seven folds. Within A and B, each panel represents a benchmark dataset. Colors indicate fixed ensemble sizes $k=1,\ldots,10$, while white boxes denote automatic selection of $k$.}\label{app:fig:bench_num_splits}
 \end{figure}

\begin{figure}[ht!]
  \centering
  \includegraphics[width=\linewidth]{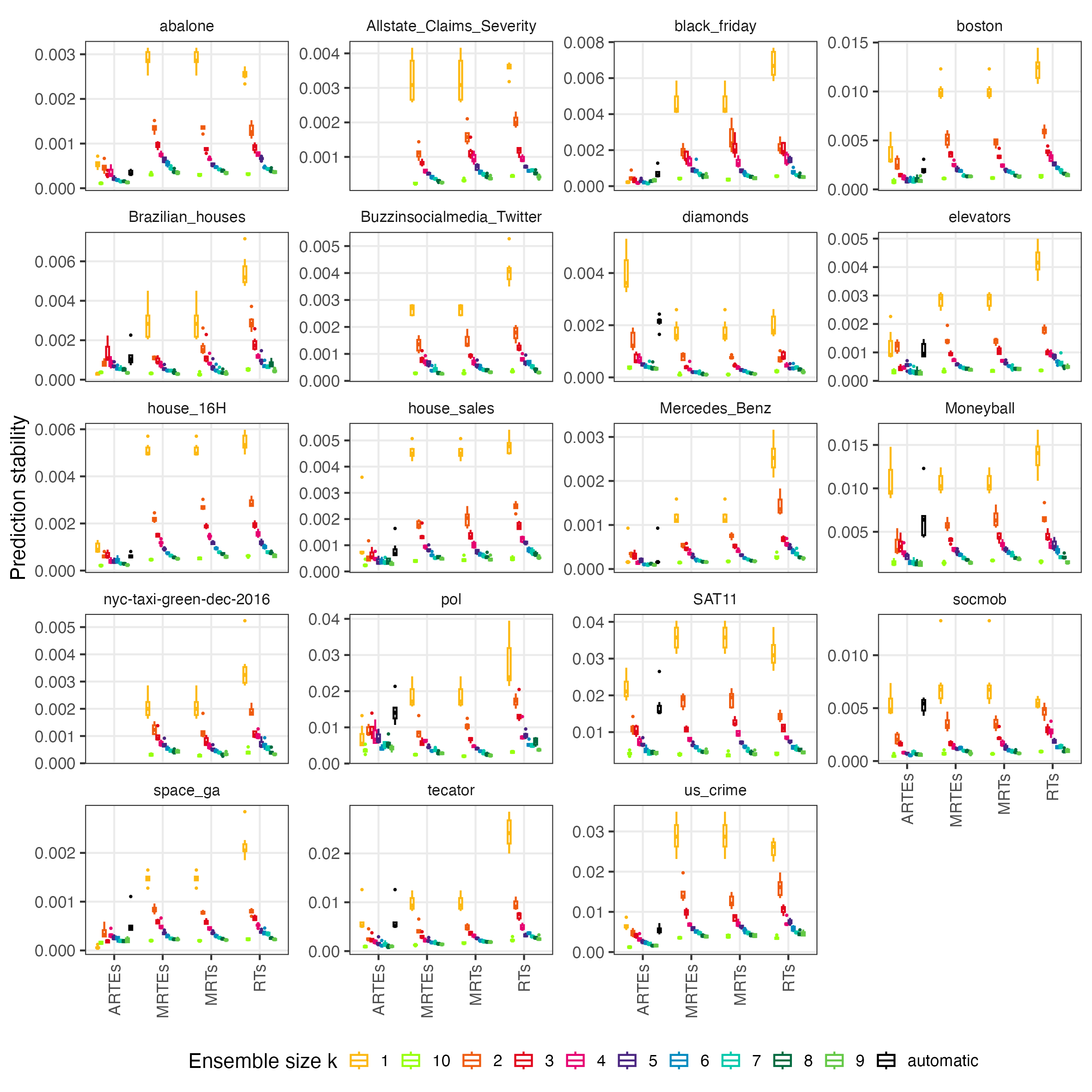}
  \caption{Prediction stability across cross-validation folds for benchmark datasets with RF out-of-bag $R^2 \geq 0.3$. Each boxplot shows the distribution of model-specific mean pairwise prediction MSE values calculated on common reference observations from the corresponding preprocessed dataset. Lower values indicate greater prediction stability. Each panel represents a dataset. Colors indicate fixed ensemble sizes $k=1,\ldots,10$, while white boxes denote automatic selection of $k$. ARTEs use Boruta variable selection. Clustering uses WSV distances.}
  \label{app:fig:bench_prediction_stability}
\end{figure}

\begin{figure}[p]
  \centering
  \includegraphics[width=\linewidth]{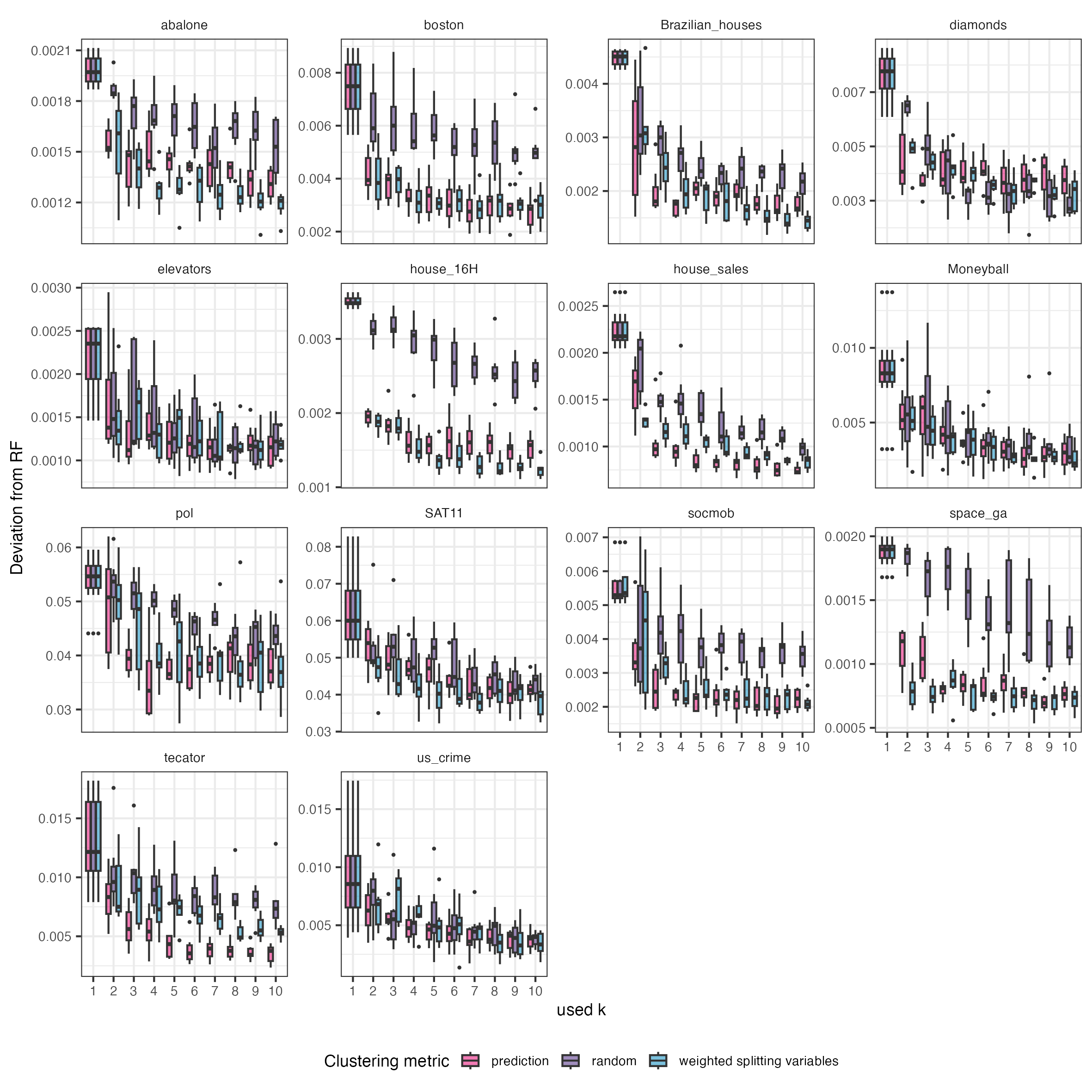}
  \caption{Sensitivity analysis comparing weighted splitting variables (WSV) based, prediction-based, and random clustering for ART ensembles at fixed ensemble sizes $k=1,\ldots,10$. Each panel represents a benchmark dataset. Boxplots summarize the MSE between ART ensemble and RF predictions on the held-out folds. Lower values indicate greater fidelity. ART construction uses WSV and Boruta variable selection in all three settings. Prediction-based clustering uses RF tree predictions on the training observations, random clustering assigns RF trees to approximately equally sized random groups.}
  \label{app:fig:bench_cluster_fidelity}
\end{figure}

\begin{figure}[p]
  \centering
  \includegraphics[width=\linewidth]{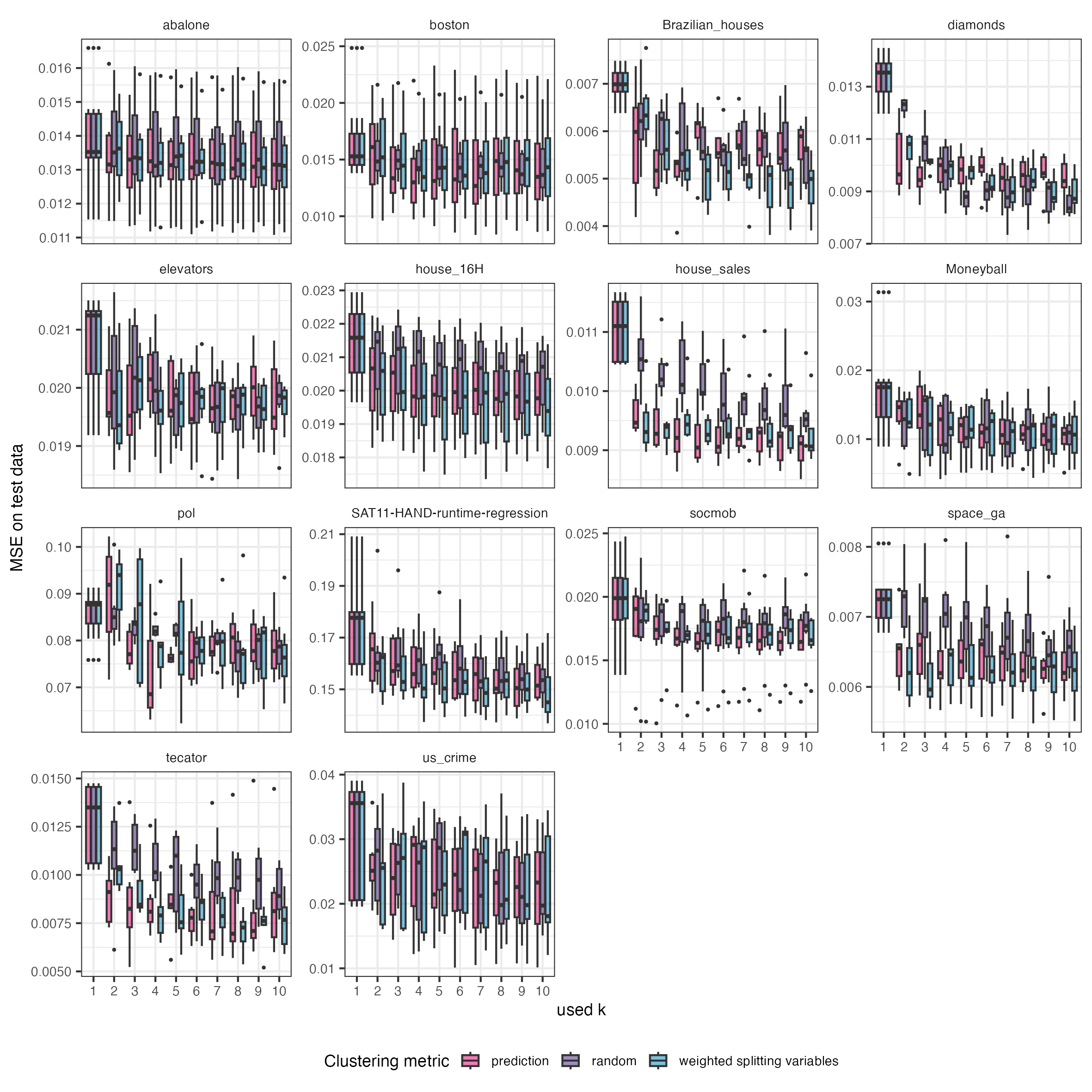}
  \caption{Test MSE for ART ensembles using weighted splitting variables (WSV) based, prediction-based, or random clustering at fixed ensemble sizes $k=1,\ldots,10$. Each panel represents a benchmark dataset. Boxplots summarize results on the held-out cross-validation folds. ART construction uses WSV and Boruta variable selection in all three settings.}
  \label{app:fig:bench_cluster_accuracy}
\end{figure}

\begin{figure}[p]
  \centering
  \includegraphics[width=\linewidth]{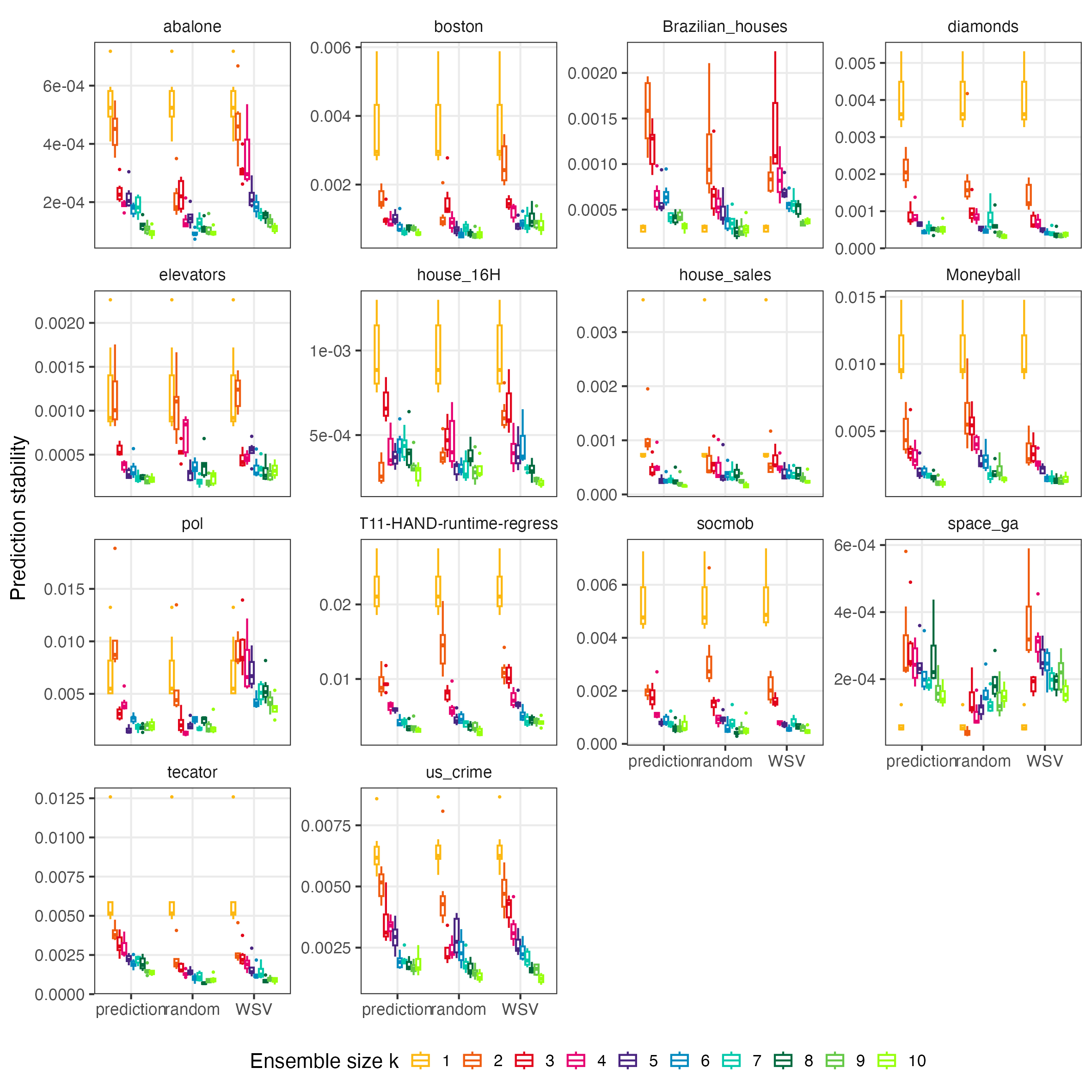}
  \caption{Prediction stability of ART ensembles using weighted splitting variables (WSV) based, prediction-based, or random clustering at fixed ensemble sizes $k=1,\ldots,10$. Each boxplot summarizes seven model-specific mean pairwise prediction MSE values across cross-validation folds, calculated on the same preprocessed reference dataset for all three clustering methods. Lower values indicate greater stability. Only dataset--ensemble-size combinations with all seven folds available for each clustering method are included. ART construction uses WSV and Boruta variable selection in all three settings.}
  \label{app:fig:bench_cluster_stability}
\end{figure}

\begin{figure}[p]
  \centering
  \includegraphics[width=\linewidth]{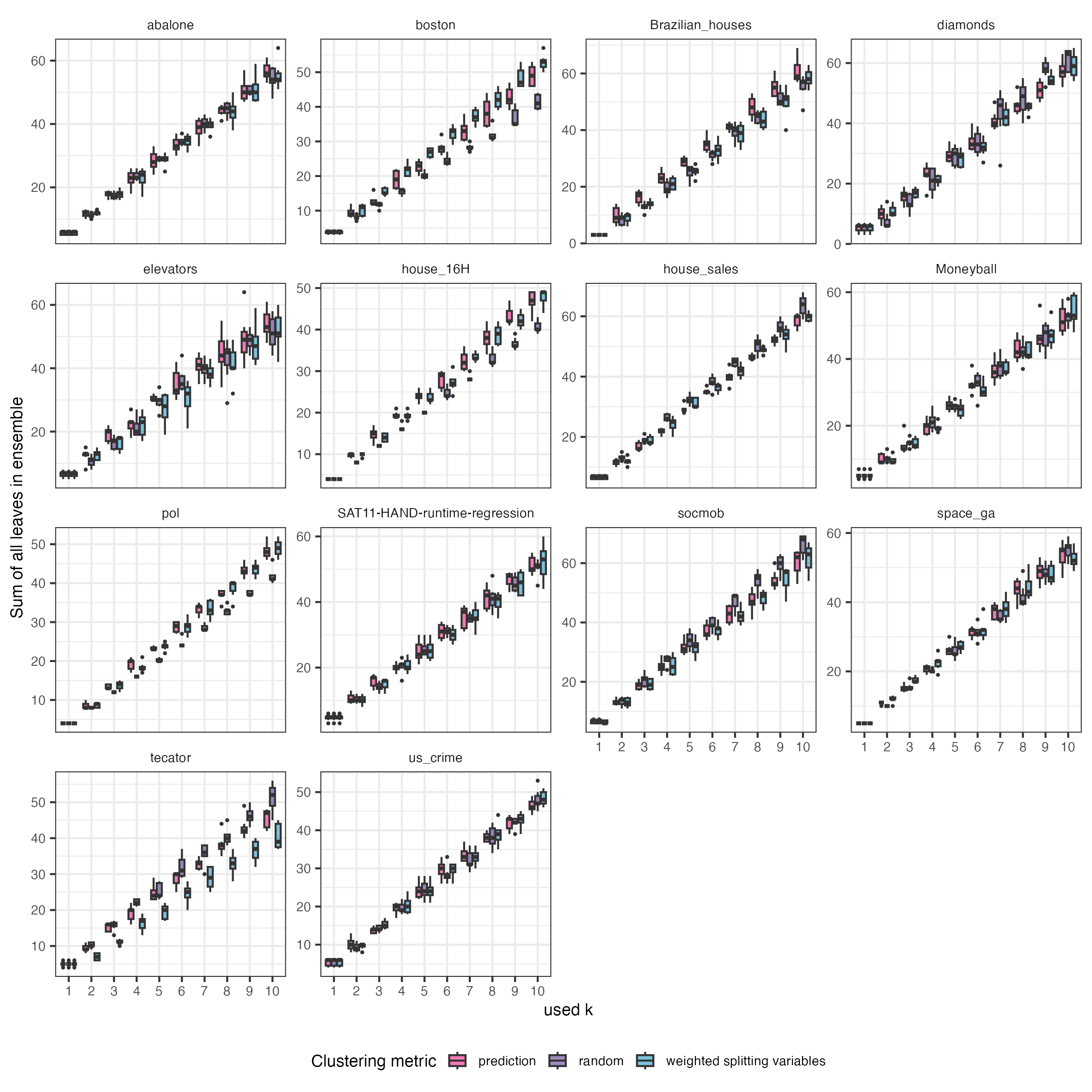}
  \caption{Total number of terminal nodes across ART ensembles using weighted splitting variables (WSV) based, prediction-based, or random clustering at fixed ensemble sizes $k=1,\ldots,10$. Each panel represents a benchmark dataset. Boxplots summarize results across cross-validation folds. ART construction uses WSV and Boruta variable selection in all three settings.}
  \label{app:fig:bench_cluster_complexity}
\end{figure}

\begin{figure}[p]
  \centering
  \includegraphics[width=\linewidth]{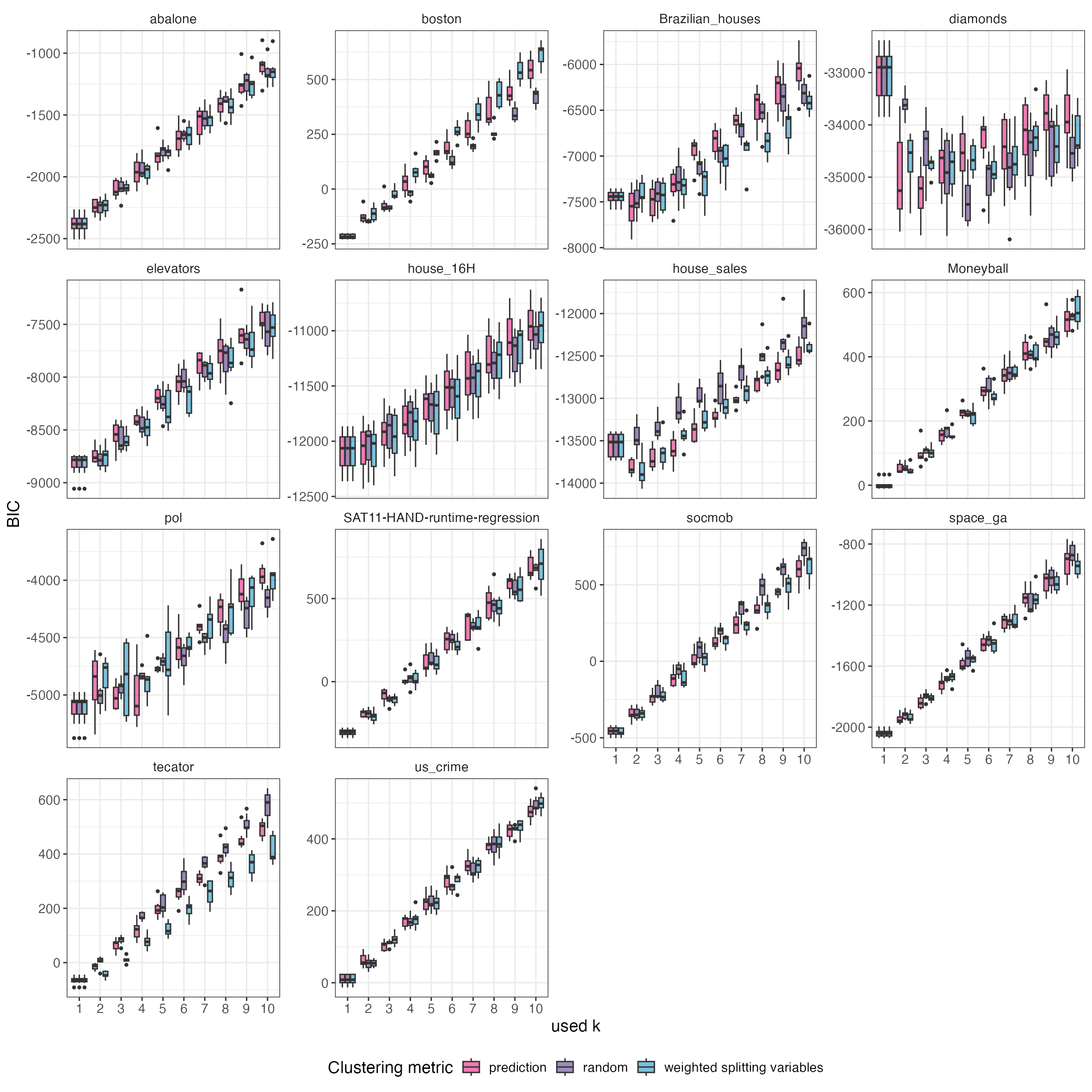}
  \caption{BIC for ART ensembles using weighted splitting variables (WSV) based, prediction-based, or random clustering at fixed ensemble sizes $k=1,\ldots,10$. Each panel represents a benchmark dataset. Boxplots summarize results across cross-validation folds. ART construction uses WSV and Boruta variable selection in all three settings.}
  \label{app:fig:bench_cluster_bic}
\end{figure}

\begin{figure}[p]
  \centering
  \includegraphics[width=\linewidth]{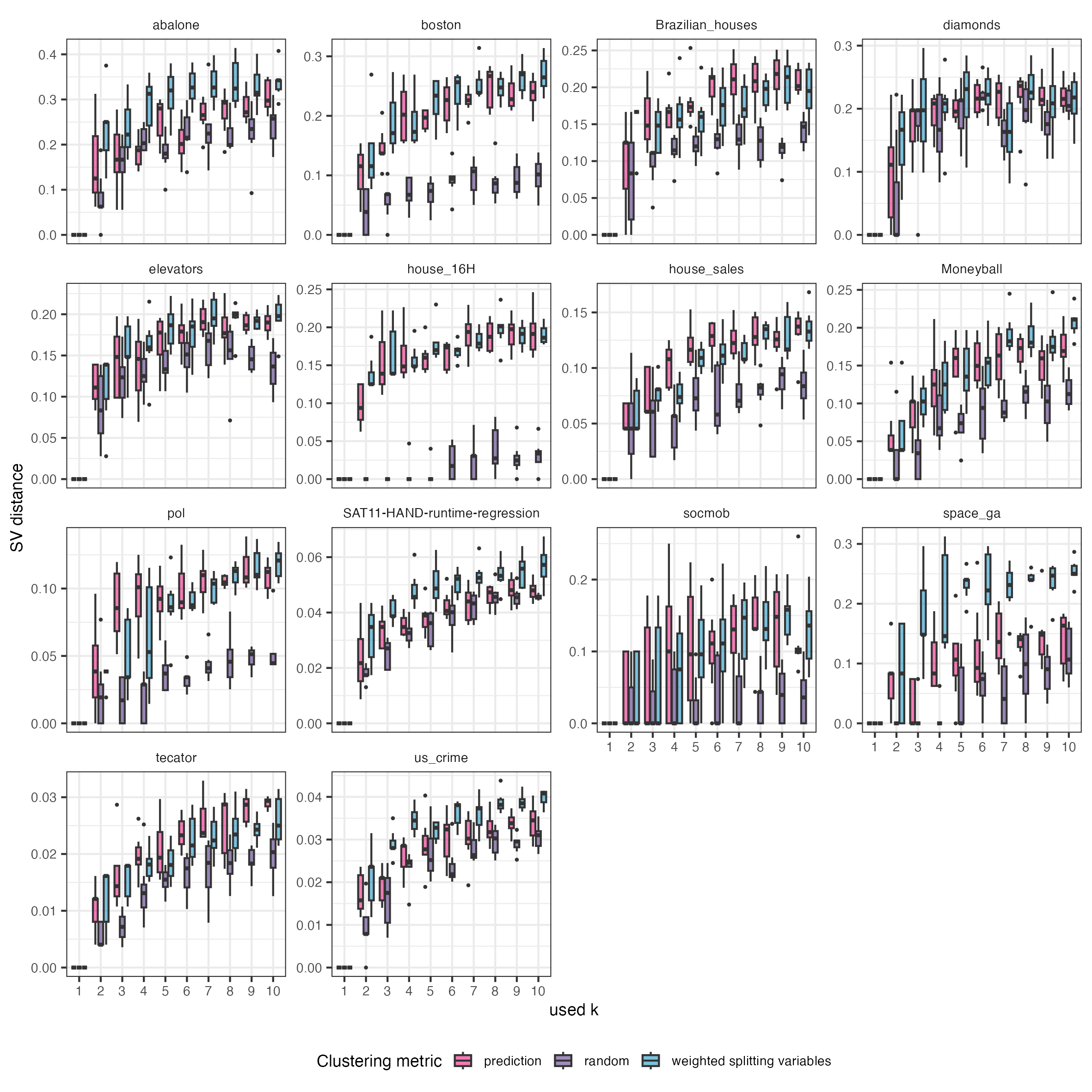}
  \caption{Mean pairwise splitting-variable (SV) distances within ART ensembles using weighted splitting variables (WSV) based, prediction-based, or random clustering. Each panel represents a benchmark dataset. Boxplots summarize results across cross-validation folds. Distances are defined for ensembles with at least two trees, the value stored for $k=1$ is a placeholder. ART construction uses WSV and Boruta variable selection in all three settings.}
  \label{app:fig:bench_cluster_sv}
\end{figure}

\begin{figure}[p]
  \centering
  \includegraphics[width=\linewidth]{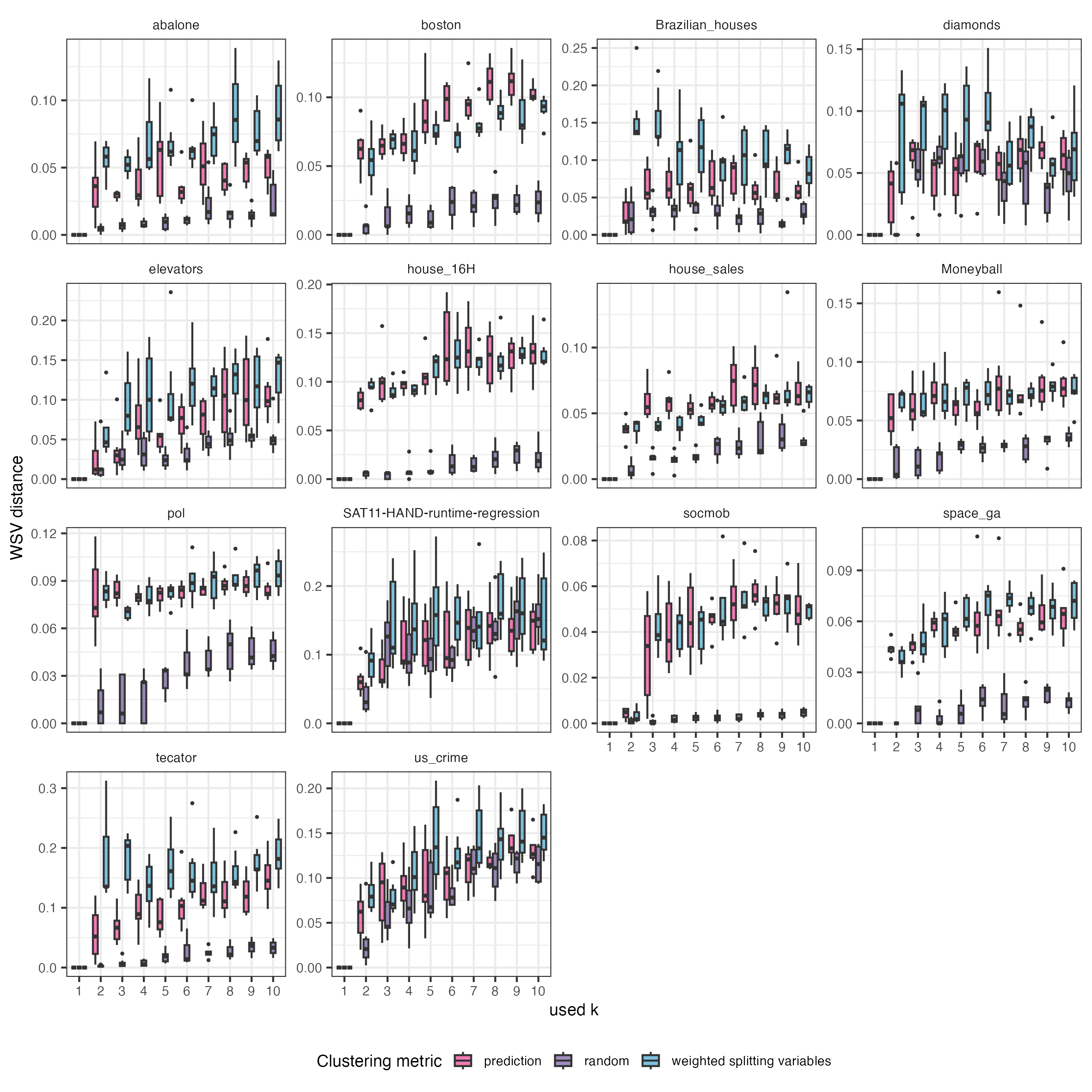}
  \caption{Mean pairwise weighted splitting-variable (WSV) distances within ART ensembles using weighted splitting variables (WSV) based, prediction-based, or random clustering. Each panel represents a benchmark dataset. Boxplots summarize results across cross-validation folds. Distances are defined for ensembles with at least two trees, the value stored for $k=1$ is a placeholder. ART construction uses WSV and Boruta variable selection in all three settings.}
  \label{app:fig:bench_cluster_wsv}
\end{figure}

\begin{figure}[p]
  \centering
  \includegraphics[width=\linewidth]{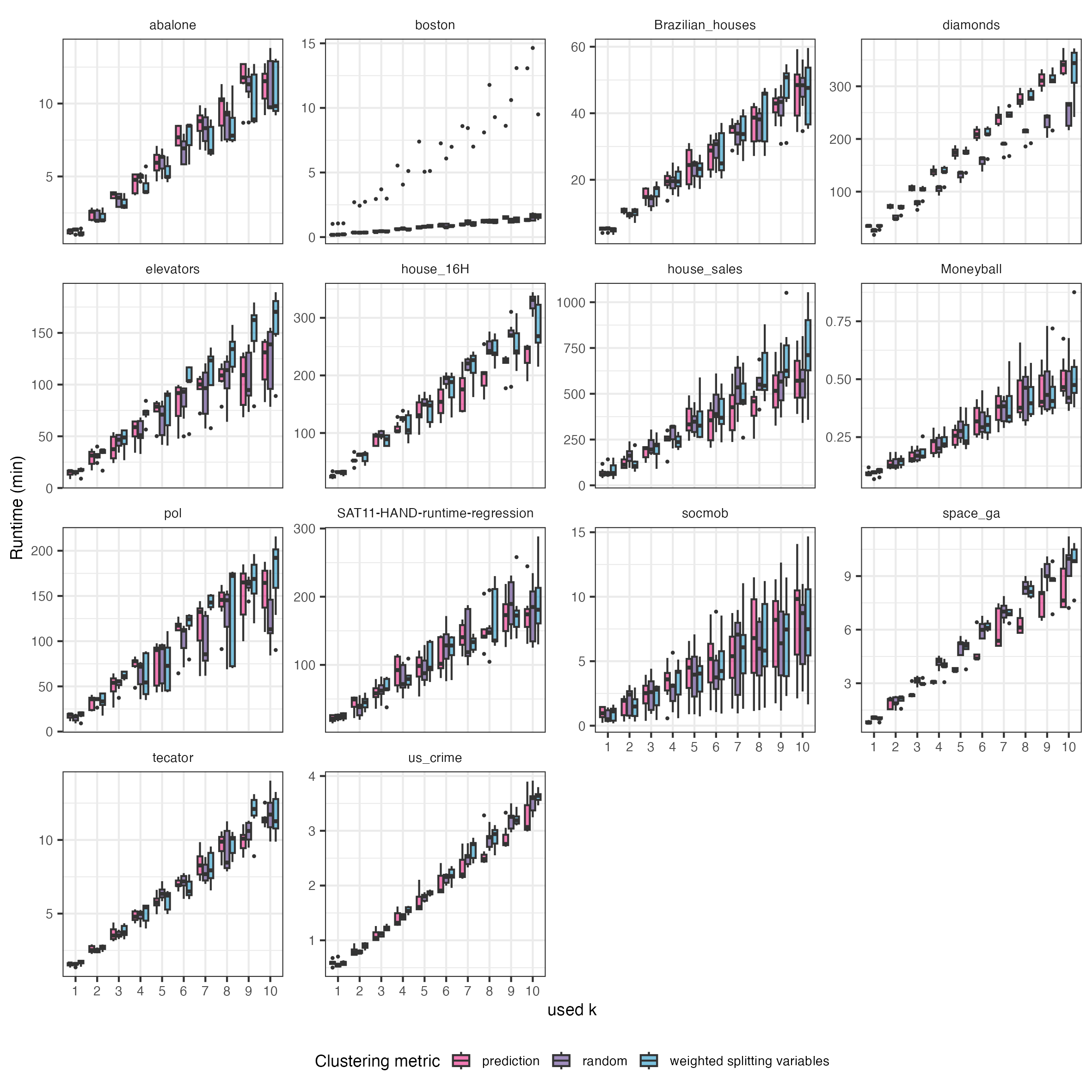}
  \caption{Runtime in minutes for constructing ART ensembles using weighted splitting variables (WSV) based, prediction-based, or random clustering at fixed ensemble sizes $k=1,\ldots,10$. Each panel represents a benchmark dataset. Boxplots summarize results across cross-validation folds. ART construction uses WSV and Boruta variable selection in all three settings.}
  \label{app:fig:bench_cluster_runtime}
\end{figure}

\FloatBarrier

\begin{table}[h]
    \centering
    \small
  
    \setlength{\tabcolsep}{3pt}
    \caption{Number of missing ART ensemble results for benchmark datasets shown in the figures with incomplete ART results. Seven cross-validation results were expected for each fixed ensemble size $k=1,\ldots,10$ and for automatic selection of $k$, giving 77 results per dataset. For data sets with incomplete ART results, only results for the other methods are shown.}
    \label{app:tab:benchmark_missing_arte}
    \begin{tabular}{p{4cm}*{11}{r}r}
      \toprule
      & \multicolumn{11}{c}{Ensemble size $k$} & \\
      \cmidrule(lr){2-12}
      Dataset
      & 1 & 2 & 3 & 4 & 5 & 6 & 7 & 8 & 9 & 10
      & Auto & Total \\
      \midrule
      Allstate\_\allowbreak Claims\_\allowbreak Severity
        & 7 & 7 & 7 & 7 & 7 & 7 & 7 & 7 & 7 & 7 & 7 & 77 \\
      black\_friday
        & 0 & 0 & 0 & 0 & 0 & 0 & 1 & 2 & 3 & 6 & 3 & 15 \\
      Buzzinsocialmedia\_\allowbreak Twitter
        & 7 & 7 & 7 & 7 & 7 & 7 & 7 & 7 & 7 & 7 & 7 & 77 \\
      Mercedes\_\allowbreak Benz\_\allowbreak Greener\_\allowbreak Manufacturing
        & 0 & 0 & 0 & 0 & 0 & 0 & 0 & 4 & 4 & 6 & 0 & 14 \\
      nyc-taxi-green-dec-2016
        & 0 & 7 & 7 & 7 & 7 & 7 & 7 & 7 & 7 & 7 & 7 & 70 \\
      \midrule
      Total missing
        & 14 & 21 & 21 & 21 & 21 & 21
        & 22 & 27 & 28 & 33 & 24 & 253 \\
      \bottomrule
    \end{tabular}
  \end{table}

\clearpage

\begin{landscape}
  \begingroup
  
  \scriptsize
  \setlength{\tabcolsep}{3pt}
  
  \begin{longtable}{p{3.6cm}rrrrrrp{1.2cm}ccp{3cm}}
  \caption{Overview of all datasets from the OpenML AutoML Benchmark Regression Collection \url{https://openml.org/search?type=benchmark&sort=tasks_included&study_type=task&id=269}. Task IDs identify OpenML tasks. The numbers of observations ($n$) and variables ($p$) are reported before and after quality control (QC). Variable counts include the outcome variable. Variable types, the OOB $R^2$ of the RF, inclusion status, and exclusion reasons are provided. Variable types are abbreviated as num. (numerical) and cat. (categorical). Inclusion refers to the benchmark evaluation irrespective of the availability of complete ART ensemble results. NA indicates unavailable information.}
  \label{app:tab:bench:data}\\
  \toprule
  Task name & Task ID &
  \shortstack{$n$\\orig.} &
  \shortstack{$p$\\orig.} &
  \shortstack{$n$\\after QC} &
  \shortstack{$p$\\after QC} &
  \shortstack[l]{Variable\\types} & $R^2$ of RF & Used & Reason, if not used\\
  \midrule
  \endfirsthead
  
  \toprule
  Task name & Task ID &
  \shortstack{$n$\\orig.} &
  \shortstack{$p$\\orig.} &
  \shortstack{$n$\\after QC} &
  \shortstack{$p$\\after QC} &
  \shortstack[l]{Variable\\types} & $R^2$ of RF & Used & Reason, if not used\\
  \midrule
  \endhead
  
  \midrule
  \multicolumn{10}{r}{\textit{Continued on next page}}\\
  \endfoot
  
  \bottomrule
  \endlastfoot
  
  diamonds & 233211 & 53940 & 10 & 53940 & 10 &
  \shortstack[l]{num.,\\cat.} & 0.895 & yes &  \\
  
  us\_crime & 359945 & 1994 & 128 & 123 & 128 &
  num. & 0.688 & yes &  \\
  
  space\_ga & 359933 & 3107 & 7 & 3106 & 7 &
  num. & 0.551 & yes &  \\
  
  wine\_quality & 359935 & 6497 & 12 & 6497 & 12 &
  num. & 0.298 & no & RF out-of-bag $R^2 < 0.3$ \\
  
  house\_16H & 359952 & 22784 & 17 & 22307 & 17 &
  num. & 0.457 & yes &  \\
  
  yprop\_4\_1 & 359940 & 8885 & 252 & 8884 & 213 &
  num. & 0.053 & no & RF out-of-bag $R^2 < 0.3$ \\
  
  sensory & 359931 & 576 & 12 & 576 & 12 &
  cat. & 0.109 & no & RF out-of-bag $R^2 < 0.3$ \\
  
  tecator & 359934 & 240 & 125 & 240 & 125 &
  num. & 0.900 & yes &  \\
  
  SAT11-HAND-runtime-regression &
  359948 & 4440 & 118 & 1725 & 116 &
  \shortstack[l]{num.,\\cat.} & 0.544 & yes &  \\
  
  abalone & 359944 & 4177 & 9 & 4170 & 9 &
  \shortstack[l]{num.,\\cat.} & 0.467 & yes &  \\
  
  MIP-2016-regression & 360945 & 1090 & 145 & 872 & 119 &
  \shortstack[l]{num.,\\cat.} & 0.030 & no & RF out-of-bag $R^2 < 0.3$ \\
  
  socmob & 359932 & 1156 & 6 & 1087 & 6 &
  \shortstack[l]{num.,\\cat.} & 0.680 & yes &  \\
  
  quake & 359930 & 2178 & 4 & 2178 & 4 &
  num. & -0.027 & no & RF out-of-bag $R^2 < 0.3$ \\
  
  boston & 359950 & 506 & 14 & 506 & 14 &
  \shortstack[l]{num.,\\cat.} & 0.799 & yes &  \\
  
  Moneyball & 167210 & 1232 & 15 & 114 & 14 &
  \shortstack[l]{num.,\\cat.} & 0.818 & yes &  \\
  
  Mercedes\_Benz\_\allowbreak Greener\_\allowbreak Manufacturing &
  233215 & 4209 & 378 & 4208 & 366 &
  \shortstack[l]{num.,\\cat.} & 0.573 & yes &  \\
  
  Santander\_\allowbreak transaction\_\allowbreak value &
  233214 & 4459 & 4993 & 4417 & 4737 &
  num. & 0.147 & no & RF out-of-bag $R^2 < 0.3$ \\
  
  Yolanda & 317614 & 400000 & 101 & 399558 & 101 &
  num. & 0.126 & no & RF out-of-bag $R^2 < 0.3$\\
  
  house\_sales & 359949 & 21613 & 23 & 21422 & 23 &
  \shortstack[l]{num.,\\cat.} & 0.660 & yes &  \\
  
  Brazilian\_houses & 359938 & 10692 & 13 & 10671 & 13 &
  \shortstack[l]{num.,\\cat.} & 0.917 & yes &  \\
  
  topo\_2\_1 & 359939 & 8885 & 267 & 8884 & 262 &
  num. & 0.031 & no & RF out-of-bag $R^2 < 0.3$ \\
  
  black\_friday & 359937 & 166821 & 10 & 166821 & 10 &
  \shortstack[l]{num.,\\cat.} & 0.367 & yes &  \\
  
  elevators & 359936 & 16599 & 19 & 16361 & 19 &
  num. & 0.434 & yes &  \\
  
  Allstate\_Claims\_\allowbreak Severity &
  233212 & 188318 & 132 & 187027 & 132 &
  \shortstack[l]{num.,\\cat.} & 0.371 & yes &  \\
  
  nyc-taxi-green-dec-2016 &
  359943 & 581835 & 19 & 577601 & 19 &
  \shortstack[l]{num.,\\cat.} & 0.477 & yes &  \\
  
  Buzzinsocialmedia\_\allowbreak Twitter &
  233213 & 583250 & 78 & 541091 & 78 &
  num. & 0.841 & yes &  \\
  
  pol & 359946 & 15000 & 49 & 15000 & 27 &
  num. & 0.885 & yes &  \\
  
  colleges & 359942 & 7063 & 45 & NA & NA &
  \shortstack[l]{num.,\\cat.} & NA & no & too many missings \\
  
  house\_prices & 359951 & 1460 & 80 & NA & NA &
  \shortstack[l]{num.,\\cat.} & NA & no & too many missings \\
  
  QSAR-TID-11 & 360932 & 5742 & 1026 & NA & NA &
  \shortstack[l]{num.,\\cat.} & NA & no & problems with \emph{OpenML} R package \\
  
  QSAR-TID-10980 & 360933 & 5766 & 1026 & NA & NA &
  \shortstack[l]{num.,\\cat.} & NA & no & problems with \emph{OpenML} R package \\
  
  Airlines\_DepDelay\_10M &
  359929 & 10000000 & 10 & 9374500 & 10 &
  \shortstack[l]{num.,\\cat.} & 0.037 & no & RF out-of-bag $R^2 < 0.3$ \\
  
  OnlineNews & 359941 & 39644 & 60 & NA & NA &
  num. & NA & no & problems with \emph{OpenML} R package \\
  
  \end{longtable}
  \endgroup
\end{landscape}

\clearpage




\end{appendices}


\bibliography{ERT_paper}

@article{breiman_random_2001,
	title = {Random forests},
	volume = {45},
	issn = {1573-0565},
	language = {en},
	urldate = {2021-05-16},
	journal = {Mach. Learn.},
	author = {Breiman, Leo},
	year = {2001},
	pages = {5--32},
}

@article{wright_ranger_2017,
	title = {ranger: a fast implementation of random forests for high dimensional data in {C}++ and {R}},
	volume = {77},
	copyright = {Copyright (c) 2017 Marvin N. Wright, Andreas Ziegler},
	issn = {1548-7660},
	shorttitle = {ranger},
	language = {en},
	urldate = {2021-05-15},
	journal = {J. Stat. Softw.},
	author = {Wright, Marvin N. and Ziegler, Andreas},
	year = {2017},
	pages = {1--17},
}

@article{kursa_feature_2010,
	title = {Feature {Selection} with the {Boruta} {Package}},
	volume = {36},
	journal = {J. Stat. Softw.},
	author = {Kursa, Miron and Rudnicki, Witold},
	month = sep,
	year = {2010},
	pages = {1--13},
}

@article{degenhardt_evaluation_2019,
	title = {Evaluation of variable selection methods for random forests and omics data sets},
	volume = {20},
	issn = {1477-4054},
	language = {eng},
	journal = {Brief. Bioinform.},
	author = {Degenhardt, Frauke and Seifert, Stephan and Szymczak, Silke},
	month = mar,
	year = {2019},
	pages = {492--503},
}

@article{seifert_surrogate_2019,
	title = {Surrogate minimal depth as an importance measure for variables in random forests},
	volume = {35},
	issn = {1367-4803},
	urldate = {2023-06-20},
	journal = {Bioinformatics},
	author = {Seifert, Stephan and Gundlach, Sven and Szymczak, Silke},
	year = {2019},
	pages = {3663--3671},
}

@article{sies_c443_2020,
	title = {C443: a methodology to see a forest for the trees},
	volume = {37},
	issn = {1432-1343},
	shorttitle = {C443},
	language = {en},
	urldate = {2024-03-21},
	journal = {J. Classif.},
	author = {Sies, Aniek and Van Mechelen, Iven},
	year = {2020},
	pages = {730--753},
}

@inproceedings{laabs_construction_2024,
	address = {Cham},
	title = {Construction of artificial most representative trees by minimizing tree-based distance measures},
	isbn = {978-3-031-63797-1},
	language = {en},
	booktitle = {Explainable {Artificial} {Intelligence}},
	publisher = {Springer Nature Switzerland},
	author = {Laabs, Björn-Hergen and Kronziel, Lea L. and König, Inke R. and Szymczak, Silke},
	editor = {Longo, Luca and Lapuschkin, Sebastian and Seifert, Christin},
	year = {2024},
	pages = {290--310},
}

@article{ishwaran_random_2008,
	title = {Random survival forests},
	volume = {2},
	journal = {Ann. Appl. Stat.},
	author = {Ishwaran, Hemant and Kogalur, Udaya B. and Blackstone, Eugene H. and Lauer, Michael S.},
	year = {2008},
	pages = {841--860},
}

@article{laabs_identification_2024,
	title = {Identification of representative trees in random forests based on a new tree-based distance measure},
	volume = {18},
	issn = {1862-5355},
	journal = {Adv. Data Anal. Classif.},
	author = {Laabs, Björn-Hergen and Westenberger, Ana and König, Inke R.},
	year = {2024},
	pages = {363--380},
}

@article{lang_batchtools_2017,
	title = {batchtools: tools for {R} to work on batch systems},
	volume = {2},
	journal = {J. Open Source Softw.},
	author = {Lang, Michel and Bischl, Bernd and Surmann, Dirk},
	year = {2017},
	pages = {135},
}

@article{banerjee_identifying_2012,
	title = {Identifying representative trees from ensembles},
	volume = {31},
	journal = {Stat. Med.},
	author = {Banerjee, Mousumi and Ding, Ying and Noone, Anne-Michelle},
	year = {2012},
	pages = {1601--1616},
}

@inproceedings{parisot_using_2013,
	address = {London, United Kingdom},
	title = {Using {Clustering} to {Improve} {Decision} {Trees} {Visualization}},
	language = {en},
	urldate = {2025-05-19},
	booktitle = {2013 17th {International} {Conference} on {Information} {Visualisation}},
	publisher = {IEEE},
	author = {Parisot, Olivier and Didry, Yoann and Tamisier, Thomas and Otjacques, Benoit},
	month = jul,
	year = {2013},
	pages = {186--191},
}

@misc{noauthor_openml_nodate,
	title = {{OpenML}},
	url = {https://www.openml.org/search?type=benchmark&sort=tasks_included&study_type=task&id=269},
	urldate = {2025-09-19},
}

@article{deng_interpreting_2019,
	title = {Interpreting tree ensembles with {inTrees}},
	volume = {7},
	issn = {2364-4168},
	language = {en},
	urldate = {2026-01-17},
	journal = {Int. J. Data Sci. Anal.},
	author = {Deng, Houtao},
	month = jun,
	year = {2019},
	pages = {277--287},
}

@article{friedman_predictive_2008,
	title = {Predictive learning via rule ensembles},
	volume = {2},
	issn = {1932-6157, 1941-7330},
	language = {en},
	urldate = {2026-03-25},
	journal = {Ann. Appl. Stat.},
	publisher = {Institute of Mathematical Statistics},
	author = {Friedman, Jerome H. and Popescu, Bogdan E.},
	month = sep,
	year = {2008},
	pages = {916--954},
}

@article{szepannek_cant_2024,
	title = {Can’t see the forest for the trees},
	volume = {51},
	issn = {1349-6964},
	language = {en},
	urldate = {2026-05-03},
	journal = {Behaviormetrika},
	author = {Szepannek, Gero and Holt, Björn-Hergen von},
	month = jan,
	year = {2024},
	pages = {411--423},
}

@inproceedings{benard_interpretable_2021,
	series = {Proceedings of {Machine} {Learning} {Research}},
	title = {Interpretable {Random} {Forests} via {Rule} {Extraction}},
	volume = {130},
	booktitle = {Proceedings of {The} 24th {International} {Conference} on {Artificial} {Intelligence} and {Statistics}},
	publisher = {PMLR},
	address = {\relax},
	author = {Bénard, Clément and Biau, Gérard and da Veiga, Sébastien and Scornet, Erwan},
	editor = {Banerjee, Arindam and Fukumizu, Kenji},
	month = apr,
	year = {2021},
	pages = {937--945},
}

@article{grubinger_evtree_2014,
	title = {evtree: {Evolutionary} {Learning} of {Globally} {Optimal} {Classification} and {Regression} {Trees} in {R}},
	volume = {61},
	copyright = {Copyright (c) 2012 Thomas Grubinger, Achim Zeileis, Karl-Peter Pfeiffer},
	issn = {1548-7660},
	shorttitle = {evtree},
	language = {en},
	urldate = {2026-10-30},
	journal = {J. Stat. Softw.},
	author = {Grubinger, Thomas and Zeileis, Achim and Pfeiffer, Karl-Peter},
	month = oct,
	year = {2014},
	pages = {1--29},
}

@article{jiang_neighborhood_2025,
	title = {A neighborhood rough sets-based ensemble method, with application to software fault prediction},
	volume = {264},
	issn = {0957-4174},
	urldate = {2026-10-31},
	journal = {Expert Syst. Appl.},
	author = {Jiang, Feng and Hu, Qiang and Yang, Zhiyong and Liu, Jinhuan and Du, Junwei},
	month = mar,
	year = {2025},
	pages = {125919},
}

@article{jiang_ensemble_2026,
	title = {An ensemble method using neighborhood granular combination entropy for software defect prediction},
	volume = {63},
	issn = {0306-4573},
	urldate = {2026-10-31},
	journal = {Inf. Process. Manag.},
	author = {Jiang, Feng and Yu, Xu and Hu, Qiang and Liu, Jinhuan and Du, Junwei},
	month = mar,
	year = {2026},
	pages = {104483},
}

@article{tak_meta_2021,
	title = {Meta fuzzy functions based feed-forward neural networks with a single hidden layer for forecasting},
	volume = {91},
	issn = {0094-9655},
	urldate = {2026-10-31},
	journal = {J. Stat. Comput. Simul.},
	publisher = {Taylor \& Francis},
	author = {Tak, Nihat},
	month = sep,
	year = {2021},
	pages = {2800--2816},
}

@article{ucan_bootstrap-based_2026,
	title = {Bootstrap-based {K}-means feature selection strategy for fuzzy regression functions},
	volume = {16},
	copyright = {2026 The Author(s)},
	issn = {2045-2322},
	language = {en},
	urldate = {2026-10-31},
	journal = {Sci. Rep.},
	publisher = {Nature Publishing Group},
	author = {Ucan, Aylin and Yildiz, Dogan and Tak, Nihat},
	month = may,
	year = {2026},
	pages = {23167},
}

@misc{bischl_openml_2017,
	title = {{OpenML} {Benchmarking} {Suites} and the {OpenML100}},
	urldate = {2026-11-03},
	publisher = {arXiv},
	author = {Bischl, Bernd and Casalicchio, Giuseppe and Feurer, Matthias and Hutter, Frank and Lang, Michel and Mantovani, Rafael G. and Rijn, Jan N. van and Vanschoren, Joaquin},
	month = aug,
	year = {2017},
	note = {arXiv:1708.03731 [stat.ML]
version: 1},
}

\end{document}